\documentclass[twoside]{article}

\usepackage[accepted]{aistats2026}

\usepackage[round]{natbib}

\usepackage{microtype}
\usepackage{graphicx}
\usepackage{booktabs} %
\usepackage{subcaption}
\usepackage[ruled,vlined]{algorithm2e}

\usepackage{tikz}
\usetikzlibrary{arrows.meta}

\usepackage{hyperref}

\usepackage{amsmath}
\usepackage{amssymb}
\usepackage{mathtools}
\usepackage{amsthm}

\usepackage[capitalize,noabbrev]{cleveref}

\theoremstyle{plain}
\newtheorem{theorem}{Theorem}[section]

\newtheorem{lemma}[theorem]{Lemma}

\theoremstyle{definition}
\newtheorem{definition}[theorem]{Definition}
\newtheorem{assumption}[theorem]{Assumption}
\theoremstyle{remark}
\newtheorem{remark}[theorem]{Remark}

\usepackage[utf8]{inputenc} %
\usepackage[T1]{fontenc}    %
\usepackage{hyperref}       %
\usepackage{url}            %
\usepackage{booktabs}       %
\usepackage{amsfonts}       %
\usepackage{nicefrac}       %
\usepackage{microtype}      %
\usepackage{xcolor}         %
\usepackage{adjustbox}

\begin{document}

\twocolumn[

\aistatstitle{Efficient Subgroup Analysis via Optimal Trees with Global Parameter Fusion
}

\aistatsauthor{ Zhongming Xie \And  Joseph Giorgio \And  Jingshen Wang }

\aistatsaddress{ University of California, Berkeley \And   University of California, Berkeley \And  University of California, Berkeley} ]

\begin{abstract}
Identifying and making statistical inferences on differential treatment effects—commonly known as subgroup analysis in clinical research—is central to precision health. Subgroup analysis allows practitioners to pinpoint populations for whom a treatment is especially beneficial or protective, thereby advancing targeted interventions. Tree-based recursive partitioning methods are widely used for subgroup analysis due to their interpretability. Nevertheless, these approaches encounter significant limitations, including suboptimal partitions induced by greedy heuristics and overfitting from locally estimated splits, especially under limited sample sizes. To address these limitations, we propose a fused optimal causal tree method that leverages mixed-integer optimization (MIO) to facilitate precise subgroup identification. Our approach ensures globally optimal partitions and introduces a parameter-fusion constraint to facilitate information sharing across related subgroups. This design substantially improves subgroup discovery accuracy and enhances statistical efficiency. We provide theoretical guarantees by rigorously establishing out-of-sample risk bounds and comparing them with those of classical tree-based methods. Empirically, our method consistently outperforms popular baselines in simulations. Finally, we demonstrate its practical utility through a case study on the Health and Aging Brain Study–Health Disparities (HABS-HD) dataset, where our approach yields clinically meaningful insights.

\end{abstract}

\section{INTRODUCTION}

\subsection{Motivation and Challenges}\label{sec:challenges}

In clinical research, investigators often aim to assess the heterogeneity of treatment effects across different patient sub-populations, commonly referred to as subgroup analysis \citep{yusuf1991analysis,rothwell2005subgroup,lipkovich2017tutorial}. Such analyses provide valuable information for clinical decision-making and future medical research, as they help identify which patients may experience beneficial or protective effects from a treatment. For example, recent trials of Lecanemab and Donanemab have shown that reducing Amyloid-beta ($A\beta$) plaques leads to measurable cognitive benefits in treating Alzheimer’s Disease (AD) \citep{van2023lecanemab,sims2023donanemab}. However, the progression of AD pathology and its impact on cognition can vary substantially across populations with distinct genetic profiles and tau levels. Studies on Lecanemab reported limited cognitive benefits among subgroups such as females, non-white participants, and APOE4 carriers \citep{van2023lecanemab}, while the Donanemab trial showed no significant effects in a prespecified cohort with high tau pathology \citep{sims2023donanemab}. These findings underscore the need for statistical methodologies capable of uncovering heterogeneity, offering insights for personalized treatment strategies.

Recursive partitioning tree-based approaches \citep{su2009subgroup,loh2015regression} are widely used in subgroup analysis in clinical research. They partition the covariate space and provide highly interpretable subgroups, facilitating the discovery of heterogeneous subpopulations. Despite their popularity, applying these methods to biomedical studies poses several challenges:

(1) \textit{Suboptimal and unstable subgroup partitions that fail to capture true heterogeneity, especially in underrepresented populations with limited sample sizes.} Clinical trials often have limited enrollment due to eligibility restrictions, high costs, and recruitment barriers, leaving some populations (e.g., racial minorities or rare biomarker groups) with very small sample sizes. With such limited data, tree-based methods for subgroup identification become especially unstable and prone to spurious splits, making it difficult to detect true heterogeneity. This instability arises because these methods rely on greedy heuristics, generating splits in isolation without considering the downstream impact of future splits in the tree \citep{bertsimas2017optimal}. As a result, the identified subgroups may misrepresent the underlying subpopulations and lead to misleading subgroup conclusions.

(2) \textit{Low-prevalence but influential covariates require borrowing information across subgroups to attain reliable insights.} Clinical outcomes may be strongly affected by covariates that are highly relevant yet rare in the study population. For example, AD is tightly linked to rare genetic risk factors such as APOE-$\varepsilon$4, which are carried by only about 5\% of the population. When the entire study sample size is already small, analyzing such covariates in isolation becomes impractical, as limited subgroup-level data increase the risk of overfitting. To obtain stable and reliable insights, it is therefore essential to borrow information (“borrow strength”) across subgroups. However, standard recursive-partitioning trees are unable to achieve this: each split is chosen based solely on local data, ignoring the global structure of the dataset and limiting the ability to detect true heterogeneity \citep{tan2022cautionary}.

\subsection{Our Solution and Contributions}\label{sec:contri}

To address these challenges and enhance the discovery of more accurate and informative subpopulations, we propose a \textit{fused optimal causal tree}, which offers two primary contributions over existing tree-based subgroup analysis methods:

\textbf{Data-efficient and globally optimal subgroup partitions achieved by mixed-integer optimization:} 
Rather than employing greedy tree-based methods, we frame the subgroup discovery problem as a comprehensive mixed integer optimization (MIO) problem, with an objective function tailored to promote heterogeneous causal discovery of subpopulations. This formulation allows us to construct the optimal causal tree by addressing a single comprehensive optimization problem (Algorithm \ref{algo:mio}), rather than decomposing it into isolated sub-problems solved greedily and resulting in sub-optimal solutions.  We also provide a theoretical analysis to demonstrate that the proposed method consistently uncovers true subgroup partitions without requiring an excessive tree depth (Theorem \ref{thm:OT}). In contrast, conventional greedy-based trees, like CART \citep{breiman1984cart}, often require infinite depth to achieve similar consistency.

\textbf{Allowing information sharing between subgroups to encourage more informative subgroup discovery and enhance statistical estimation efficiency:} To facilitate between-subgroup information sharing, we impose a parameter-fusion constraint on the leaf nodes. This encourages groups with similar effects to borrow strength, reducing overfitting in small samples. Greedy tree algorithms cannot handle such global constraints because they optimize splits one node at a time. Our mixed-integer programming formulation embeds fusion directly into the full optimization, letting the entire dataset inform subgroup boundaries and treatment-effect estimates. As indicated by our simulations in Section \ref{sec:simulation}, this fusion constraint not only improves the accuracy of subgroup identification but also produces more efficient causal estimates. With this method, we can provide more reliable insights into which subpopulations are most likely to benefit from specific treatments.

\subsection{Related Literature}

Regression-based subgroup identification methods have gained popularity in subgroup analysis due to their ability to model heterogeneity through interaction terms. Notably, \cite{tian2014simple} and \cite{shen2015inference} incorporate interactions between the treatment and pre-specified subgroup indicators within regression models. By adopting a variable selection perspective, these methods aim to identify significant interactions that indicate differential treatment effects across subgroups. However, a potential limitation is that subgroup indicators often cannot be pre-specified in practice. This constraint renders regression-based methods more restrictive compared to tree-based approaches, which can automatically discover relevant subgroups from the data without prior specification.

\cite{su2009subgroup}, \cite{athey2016recursive}, \cite{wager2018estimation}, \cite{hill2011}, \cite{zeldow2019semiparametricBART}, and \cite{hahn2020bart} develop methods that identify subgroups by recursively partitioning the data based on baseline confounders. These tree-based models are adept at capturing complex interactions and nonlinear relationships inherent in heterogeneous treatment effects. \cite{ting2023estimating} further extends the Bayesian Additive Regression Trees (BART) method to identify heterogeneous causal mediation effects involving a single mediator, showcasing the versatility of tree-based methods in causal inference. Despite their strengths, classical tree-based approaches often rely on heuristics for making split decisions in isolation, which may not yield globally optimal solutions. Similarly, clustering algorithms like \(k\)-means and Gaussian mixture models have been used to first identify subgroups and then estimate causal parameters \cite{xue2022heterogeneous}, but these methods also face challenges in achieving global optimality.

Recent literature has focused on enhancing the efficiency and optimality of classification and regression trees by employing advanced optimization techniques. Dynamic programming methods have been utilized by \cite{demirovic2022murtree}, \cite{lin2020generalized}, \cite{van2022fair}, and \cite{bos2024piecewise} to solve for optimal trees more effectively. Additionally, \cite{mazumder2022quant} apply the Branch-and-Bound method to find globally optimal tree structures. While these approaches improve the global optimality of tree-based models, they do not consider parameter fusion, which is crucial for leveraging shared information across different parts of the model. Moreover, none of these works specifically addresses causal subgroup identification. This gap highlights the need for methods that can efficiently construct globally optimal trees while incorporating parameter fusion constraints to enhance subgroup identification and treatment effect estimation.

\cite{huang2025distilling} study heterogeneous treatment effect estimation via distillation, aggregating noisy signals into stable and interpretable subgroups and highlighting the stability–efficiency tradeoff in subgroup discovery. Their focus on stabilizing heterogeneous effect estimation is closely aligned with ours, but differs in that we embed this principle directly into tree construction via mixed-integer optimization rather than relying on post-hoc distillation. Distillation and our tree construction can be viewed as complementary mechanisms for promoting stability, and may be naturally integrated within a unified framework.

\section{PROBLEM SETUP} \label{sec:prelim}

We consider the potential outcomes framework \citep{rubin2005causal} and a setup with \( n \) subjects such that \( (\boldsymbol{X}_i, T_i, Y_i(0), Y_i(1)) \sim_{i.i.d.} \mathbb{P}_0 \) for \( i = 1, \ldots, n \). Here, \( (Y_i(0), Y_i(1)) \) represents the potential outcomes for subject \( i \). $\boldsymbol{X}_i \in \mathbb{R}^d$ represents the $d$-dimensional covariates that potentially capture  heterogeneity. $T$ is a binary variable that can be intervened on. For each subject $i$, we observe $(\boldsymbol{X}_i, T_i, Y_i)$, where $Y_i=T_i\cdot Y_i(1)+(1-T_i)\cdot Y_i(0)$. 
  
We assume there exists $M^*$ unknown heterogeneous subpopulations $\mathcal{A}_1^*, \ldots, \mathcal{A}_{M^*}^*$ and let $\Pi^* = \{\mathcal{A}_m^*\}_{m=1}^{M^*}$ denote the collection of these subpopulations. Then we model the heterogeneity using the following linear structural equation model:
\begin{equation}\label{regression}
    \begin{aligned}
    & Y_i=\sum_{m=1}^{M^*} \mathrm{1}_{(\boldsymbol{X}_i\in \mathcal{A}_m^*)}\cdot f_m(\boldsymbol{X}_i, T_i)+\epsilon,
    \\& f_m(\boldsymbol{X}_i, T_i)=\delta_m^*+\mu_m^*\cdot T_i+\boldsymbol{\alpha}_m^{*\top}\cdot \boldsymbol{X}_i+\boldsymbol{\beta}_m^{*\top}\cdot T_i\boldsymbol{X}_i,
\end{aligned}
\end{equation}
where $\epsilon$ is independent of $(T_i, \boldsymbol{X}_i)$ with mean 0. This model includes both main effects and treatment-covariate interactions, allowing for flexible characterization of heterogeneity.  For notational convenience, we let $\boldsymbol{Z}_i = [1, T_i, \boldsymbol{X}_i^\top, T_i \boldsymbol{X}_i^\top]^\top$ and $\boldsymbol{\gamma}_m^* = (\delta_m^*, \mu_m^*, \boldsymbol{\alpha}_m^{*\top}, \boldsymbol{\beta}_m^{*\top})^{\top} \in \mathbb{R}^{2d+2}$.
In addition, to encourage information sharing between different subgroups,  we allow the number of unique values in the vector $\{\gamma^*_{1j},\ldots,\gamma^*_{M^*j} \}$ to be less than $M^*$. 

The above model provides a general framework for gaining deeper insights into studying heterogeneity in clinical trials. By capturing how baseline covariates and their interactions with treatment influence outcomes across subpopulations, it facilitates more interpretable subgroup analyses and supports downstream causal investigations.

Our downstream causal goal is to estimate the subgroup average treatment effect (SATE), which can be defined as:
\begin{align*}\tau(\boldsymbol{X}_i,\Pi^*)&=\sum_{m=1}^{M^*}\mathrm{1}_{(\boldsymbol{X}_i\in \mathcal{A}_m^*)}\mathbb{E}[Y_i(1)-Y_i(0)|\boldsymbol{X}_i\in \mathcal{A}_m^*]
\end{align*}
for subject $i$. However, directly estimating $\tau(\boldsymbol{X}_i,\Pi^*)$ is not straightforward because potential outcomes are subject to missingness. To address this, we impose the standard causal identification assumptions:
\begin{assumption}[Consistency]\label{assum1}
$$Y_i=T_i\cdot Y_i(1)+(1-T_i)\cdot Y_i(0).$$
\end{assumption}
\begin{assumption}[Unconfoundedness]\label{assum2}
$$(Y_i(0),Y_i(1)) \perp T_i \mid \boldsymbol{X}_i.$$
\end{assumption}

Under Assumptions \ref{assum1} and \ref{assum2}, the SATE is identifiable in the sense that
\begin{align*}
\tau(\boldsymbol{X}_i,\Pi^*)=\sum_{m=1}^{M^*}\mathrm{1}_{(\boldsymbol{X}_i\in \mathcal{A}_m^*)}(\eta_{1}(\boldsymbol{X}_i,\mathcal{A}_m^*)-\eta_{0}(\boldsymbol{X}_i,\mathcal{A}_m^*)),
\end{align*}
where $\eta_{t}(\boldsymbol{X}_i,\mathcal{A}_m^*)=\mathbb{E}[Y_i|\boldsymbol{X}_i\in \mathcal{A}_m^*, T_i=t]$.

\section{METHODOLOGY}\label{sec:method}
\subsection{Fused and globally optimal causal tree to discover heterogeneity}

To identify heterogeneous subpopulations and conduct downstream causal analysis of the effects of treatments on the outcome, we aim to estimate the true subgroup partition $\Pi^*$ and the corresponding parameters $\{\boldsymbol{\gamma}_m^*\}_{m=1}^{M^*}$ in model \eqref{regression}. To efficiently discover informative and interpretable subgroups, we propose estimating $\Pi^*$ by partitioning the covariate space via a learned hierarchical structure, analogous to recursive partitioning tree-based methods.

To be specific, we formulate the following optimization problem:
\begin{equation}\label{optimization}
    \begin{aligned}
&\min_{ \boldsymbol{\{\gamma}_m\}_{m=1}^{M(\Pi)} ,\Pi\in \mathcal{T}} \sum_{i=1}^n \left\| Y_i - \sum_{m=1}^{M(\Pi)} \mathrm{1}_{(\boldsymbol{X}_i \in \mathcal{A}_m)} \cdot \boldsymbol{\gamma}_m^{\top} \boldsymbol{Z}_i \right\|_2^2,\\
&\quad \text{s.t.} \  \gamma_{l_1,j} =  \gamma_{l_2, j}, \text{ for some } l_1\neq l_2 \text{ and some } j.
\end{aligned}
\end{equation}
where $\mathcal{T}$ denotes the collection of all possible hierarchical partitions and $M(\Pi)$ is the number of subgroups in the partitions $\Pi$. The objective function in \eqref{optimization} is designed to facilitate heterogeneous causal discovery, while the parameter fusion constraint enables information sharing across subgroups to maintain statistical efficiency.

The parameter fusion constraint here does not enforce full homogeneity across all leaf nodes but instead allows selective homogeneity across specific covariates. This flexibility accommodates both fully agnostic fusion and fusion guided by prior knowledge. For instance, if a rare genetic variant has sparse representation within certain subgroups, we may enforce parameter fusion on its coefficient to stabilize estimation.  By incorporating this constraint, we enhance efficiency for homogeneous effects, improving both subgroup identification accuracy and parameter estimation precision.

The solution to this optimization problem yields the estimated partition $\hat{\Pi} = \{\hat{\mathcal{A}}_m\}_{m=1}^{\hat{M}(\hat{\Pi})}$, where $\hat{M}(\hat{\Pi})$ is the number of identified subgroups, and the corresponding regression parameters $\{\hat{\boldsymbol{\gamma}}_m\}_{m=1}^{\hat{M}(\hat{\Pi})}$ characterize the learned structural relationships within each subgroup.

\textbf{Challenges:} Directly solving the optimization problem in \eqref{optimization} is hard. Existing recursive partitioning tree-based methods employ a greedy top-down approach to obtain a sub-optimal result (See Appendix \ref{app:detail on tree}). As explained in the introduction, this strategy presents significant limitations in the context of clinical trials with small sample sizes. In such settings, greedy methods lack guarantees of global optimality and are especially prone to generating unstable and inaccurate subgroup partitions. Such inaccurate subgroup discovery can affect downstream clinical interventions in clinical trials. Moreover, the parameter fusion constraints for subgroup information sharing are difficult to implement in the existing tree-based approaches, as they find different splits in isolation and fail to incorporate a global constraint across different partitions. In response to these challenges, we present a novel formulation of tree-based sample space partition problems
using modern mixed-integer optimization (MIO) (See Appendix \ref{app:mio} for more backgrounds) techniques presented in Algorithm \ref{algo:mio}. The adoption of MIO offers the additional advantage of seamlessly integrating the tree structure and parameter fusion constraints of Problem \eqref{optimization} into a single unified optimization problem.
 
\subsection{Solving fused optimal causal tree based on mixed integer optimization}

In this section, we propose a fused optimal causal tree framework that directly solves the optimization problem in \eqref{optimization}, yielding globally optimal subgroup partitions while allowing information sharing across subgroups. We achieve this by reformulating the problem as a mixed-integer optimization problem in Algorithm \ref{algo:mio}, which can be solved using the state-of-the-art mathematical optimization solver \citep{gurobi}. MIO offers a principled alternative by enabling the global optimization of subgroup partitions under complex structural constraints and parameter fusion constraints. This opens up new opportunities to improve both the accuracy and efficiency of subgroup analyses, particularly in settings with limited samples and partially shared effect structures. Now we present the use of MIO in Algorithm \ref{algo:mio} in detail. To save space, we put the detailed explanations of the notations in Appendix \ref{sec:notation}. We also provide a depth-2 tree to help visualize the role of each variable in Algorithm \ref{algo:mio}.

\textbf{MIO neatly represents the tree with binary indicator variables that the algorithm optimizes directly.} Let $\mathcal{T}_L$ denote the set of leaf nodes in the hierarchical partition. We reformulate the loss function using binary indicators $\{z_{it'}\}_{i \in [n],\ t' \in \mathcal{T}_L}$, where $z_{it'} = 1$ if the $i$-th subject is assigned to leaf node $t'$, and $z_{it'} = 0$ otherwise. This representation allows the objective function to be expressed as:
$$
L_n\big(\mathcal{T}_L, \{\boldsymbol{\gamma}_{t'}\}_{t' \in \mathcal{T}_L} \big) = \sum_{i \in [n]} \left\| Y_i - \sum_{t' \in \mathcal{T}_L} z_{it'} \cdot \boldsymbol{\gamma}_{t'}^\top \boldsymbol{Z}_i \right\|_2^2.
$$
To formulate $\Pi$, the subgroup partition that follows a hierarchical tree structure, we introduce several constraints, which can be divided into the following two parts.

\textbf{MIO seamlessly incorporates subgroup membership constraint.} The first part models the subgroup membership of the subjects. The first constraint ensures that each subject $i$ in a leaf node $t_0$ (where $z_{it_0} = 1$) follows a homogeneous linear regression model. The second constraint ensures that each subject belongs to exactly one leaf node. The third constraint identifies empty leaf nodes ($l_{t'} = 0$) by enforcing $z_{it'} = 0$ for all $i$. The fourth constraint requires that non-empty leaf nodes ($l_{t'} = 1$) contain at least $N_{\min}$ subjects.

\textbf{MIO precisely and flexibly represents tree structures.} The second set of constraints enforces the hierarchical structure of the partitions. We let $\mathcal{T}_B$ denote the branch nodes that partition the sample space. The first two inequalities define the tree splits using binary vectors $\{\boldsymbol{a}_m\}_{m \in \mathcal{T}_B}$ and continuous variables $\{b_m\}_{m \in \mathcal{T}_B}$. Subjects in a leaf node $t'$ are characterized by all the splits in the ancestor nodes of $t'$, with $\mathbf{a}_m^\top \mathbf{X}_i < b_m$ for left branches and $\mathbf{a}_m^\top \mathbf{X}_i \geq b_m$ for right branches. The third and fourth constraints allow branch nodes to remain unsplit ($d_{t'} = 0$) or enforce splits on a single variable for branch nodes that are split ($d_{t'} = 1$). The final constraint ensures the hierarchical structure by preventing a branch node from splitting unless its parent node has already split.

\textbf{MIO naturally accommodates parameter fusion constraints, enabling effective information sharing across subgroups.} In addition to these, we need to incorporate the parameter fusion constraint to facilitate information sharing across subgroups. To add this, we augment the objective function with an $L_0$-fusion penalty $\lambda \cdot \sum_{j =[2d+2] ,\  t_1, t_2 \in \mathcal{T}_L;\  t_1 \neq t_2} r_{jt_1t_2}$ such that
\begin{align*}
    &(\gamma_{t_1j} - \gamma_{t_2j}) \cdot (1 - r_{jt_1t_2}) = 0, 
    \\& r_{jt_1t_2} \in \{0, 1\},\  \forall j \in [2d+2], \ t_1, t_2 \in \mathcal{T}_L \text{ and } t_1 \neq t_2.
\end{align*}
Here, the parameter $\lambda$ controls the level of parameter fusion, with larger values encouraging greater homogeneity and smaller values allowing for more heterogeneity. The introduction of this fusion penalty significantly improves sample efficiency by reducing the dimensionality of the parameter space. Importantly, it can alter the learned tree structure (subgroups)—a capability not present in existing tree-based methods. An alternative is to encourage homogeneity by penalizing 
$L_2$ differences between subgroup parameters rather than enforcing exact equality. Such an $L_2$-penalty can be seamlessly incorporated into our mixed-integer framework if one prefers a soft notion of sharing. Our choice of an 
$L_0$-fusion constraint is deliberate: it yields exactly shared coefficients across subgroups, which produces a more interpretable representation of effect patterns and more aggressively borrows strength when partial shrinkage may still leave parameters weakly identified.

\textbf{Tuning parameter selection:} To tune the optimal value of $\lambda$, we employ the Bayesian Information Criterion (BIC), which is widely used in model selection \citep{schwarz1978estimating}. We select $\lambda$ that minimizes the following BIC expression:
$\text{BIC}(\lambda) = n \cdot \log (L_n\big(\hat{\mathcal{T}}_L, \{ \hat{\boldsymbol{\gamma}}_t\}_{t \in \hat{\mathcal{T}}_L}\big)/n) + \text{df}(\lambda) \cdot \log(n),
$
where $\text{df}(\lambda) = \sum_{j=1}^{2d+2} \#\{\text{Distinct } \hat{\gamma}_{tj} \text{ for } t \in \hat{\mathcal{T}}_L\}$ is the number of distinct parameters across subgroups. Typically, the choice of $\lambda$ depends on the sample size $n$. To eliminate this dependence and simplify the implementation, we normalize the loss function by $\hat{L}(\mathcal{D}_n)$, where $\hat{L}(\mathcal{D}_n)$ is the loss from fitting a single linear regression model to the entire training dataset $\mathcal{D}_n$. In our implementation, we use a grid search to select the optimal $\lambda$ by minimizing $\text{BIC}(\lambda)$. Specifically, we compute:$
\hat{\lambda}_{BIC} = \arg\min_{\lambda \in \Lambda} \text{BIC}(\lambda),$
Where $\Lambda$ is the predefined grid of candidate values for $\lambda$. We then use the corresponding $\hat{\mathcal{T}}_L, \{ \hat{\boldsymbol{\gamma}}_t\}_{t \in \hat{\mathcal{T}}_L}$ for the selected $\hat{\lambda}_{BIC}$ as the final solution.

To help readers better understand our Algorithm \ref{algo:mio}, here we use a depth-2 tree for illustration.

\textbf{Depth-2 tree structure.}
\begin{center}
\begin{tikzpicture}[font=\small, >=Stealth]
    \node[circle, draw, minimum size=8mm] (n1) at (0,0) {$1$};
    \node[circle, draw, minimum size=8mm] (n2) at (-2.5,-1) {$2$};
    \node[circle, draw, minimum size=8mm] (n3) at (2.5,-1) {$3$};

    \node[rectangle, draw, rounded corners, minimum width=8mm, minimum height=6mm] (n4) at (-3.5,-3) {$4$};
    \node[rectangle, draw, rounded corners, minimum width=8mm, minimum height=6mm] (n5) at (-1.5,-3) {$5$};
    \node[rectangle, draw, rounded corners, minimum width=8mm, minimum height=6mm] (n6) at (1.5,-3) {$6$};
    \node[rectangle, draw, rounded corners, minimum width=8mm, minimum height=6mm] (n7) at (3.5,-3) {$7$};

    \draw[very thick] (n1) -- node[midway, above, sloped] {$\mathbf{a}_1^\top \mathbf{X}_i < b_1$} (n2);
    \draw (n1) -- node[midway, above, sloped] {$\mathbf{a}_1^\top \mathbf{X}_i \ge b_1$} (n3);

    \draw (n2) -- node[midway, above, sloped] {$\mathbf{a}_2^\top \mathbf{X}_i < b_2$} (n4);
    \draw[very thick] (n2) -- node[midway, above, sloped] {$\mathbf{a}_2^\top \mathbf{X}_i \ge b_2$} (n5);

    \draw (n3) -- node[midway, above, sloped] {$\mathbf{a}_3^\top \mathbf{X}_i < b_3$} (n6);
    \draw (n3) -- node[midway, above, sloped] {$\mathbf{a}_3^\top \mathbf{X}_i \ge b_3$} (n7);
\end{tikzpicture}
\end{center}

\textbf{How to read this tree.}
Each subject \(i\) starts at the root node \(1\) and is routed downward by split rules until reaching one leaf in 
\(\mathcal{T}_L=\{4,5,6,7\}\). For example, the bold path corresponds to
\[
1 \xrightarrow{\text{left}} 2 \xrightarrow{\text{right}} 5,
\]
which implies \(
\mathbf{a}_1^\top \mathbf{X}_i < b_1 \text{ and } 
\mathbf{a}_2^\top \mathbf{X}_i \ge b_2.
\)

\textbf{Node sets and parent relation.}
\[
\mathcal{T}_B=\{1,2,3\}, \quad \mathcal{T}_L=\{4,5,6,7\}, \quad 
\texttt{p}(2)=\texttt{p}(3)=1.
\]

\textbf{Path notation.}
For leaf \(5\), the path is encoded as
\[
\mathcal{L}(5)=\{1\}, \qquad \mathcal{R}(5)=\{2\},
\]
meaning left at node \(1\) and right at node \(2\).

\textbf{Subgroup membership variables.}
For the above path, \(z_{i5}=1\) and all other \(z_{it'}=0\).

\textbf{Tree splits.}
Each branch node \(t' \in \mathcal{T}_B\) uses \((\mathbf{a}_{t'}, b_{t'})\) to define a split:
\[
\mathbf{a}_{t'}^\top \mathbf{X}_i < b_{t'} \ (\text{left}), 
\quad 
\mathbf{a}_{t'}^\top \mathbf{X}_i \ge b_{t'} \ (\text{right}).
\]

\textbf{Leaf models.}
Each leaf \(t'\) has parameters \(\boldsymbol{\gamma}_{t'}\), and the fitted value is $\sum_{t' \in \mathcal{T}_L} z_{it'} \, \boldsymbol{\gamma}_{t'}^\top \boldsymbol{Z}_i.$

\textbf{Parameter fusion.}
Binary variables \(r_{j t_1 t_2}\) determine whether coefficients are shared:
\[
r_{j t_1 t_2}=0 \;\Rightarrow\; \gamma_{t_1 j}=\gamma_{t_2 j}.
\]
For example, \(r_{j,4,5}=0\) enforces information sharing between leaves \(4\) and \(5\) for variable $j$.

\begin{algorithm}[h]
\small
\caption{Fused Optimal Causal Tree}
\label{algo:mio}
\vspace{0.3em}
\textbf{Optimization objective}:
$\min_{\Theta \in \mathbf{\Theta}}  L_n\big(\mathcal{T}_L, \{\boldsymbol{\gamma}_{t'}\}_{ t' \in \mathcal{T}_L} \big)/\hat{L}(\mathcal{D}_n)+\lambda \cdot \sum_{j =[2d+2] ,\  t_1, t_2 \in \mathcal{T}_L;\  t_1 \neq t_2} r_{jt_1t_2}$
\vspace{0.1em}
\hrule
  \vspace{0.5em}
\textbf{Constraints for the subgroup membership}:
\begin{align*}
    & \big( \sum_{t' \in \mathcal{T}_L} z_{it'} \cdot \boldsymbol{\gamma}_{t'}^\top \boldsymbol{Z}_i - \boldsymbol{\gamma}_{t_0}^\top \boldsymbol{Z}_i \big) \cdot z_{it_0} = 0, \forall i \in [n], t_0 \in \mathcal{T}_L;  \\
    & \sum_{t' \in \mathcal{T}_L} z_{i t'} = 1, \quad \forall i \in [n]; \\
    & z_{i t'} \leq l_{t'}, \quad \forall i \in [n],\quad  t' \in \mathcal{T}_L; \\
    & \sum_{i=1}^{n} z_{i t'} \geq N_{\min} l_{t'}, \quad \forall t' \in \mathcal{T}_L; \\
    & z_{i t'}, l_{t'} \in \{0,1\}, \quad \forall i \in [n], t' \in \mathcal{T}_L; 
\end{align*}
  \hrule
  \vspace{0.3em}
\textbf{Constraints for the tree structure}:
\begin{align*}
    & \mathbf{a}_m^{\top}\mathbf{X}_i \geq b_m - (1 - z_{it'}), \quad \forall i \in [n], t' \in \mathcal{T}_L, m \in \mathcal{R}(t'); \\
    & \mathbf{a}_m^{\top}\mathbf{X}_i < b_m + (1 - z_{it'}), \quad \forall i \in [n], t' \in \mathcal{T}_L, m \in \mathcal{L}(t'); \\
    & \sum_{k=1}^{d} a_{t'k} = d_{t'}, \quad \forall t' \in \mathcal{T}_B; \\
    & 0 \leq b_{t'} \leq d_{t'}, \quad \forall t' \in \mathcal{T}_B; \\
    & d_{t'} \leq d_{\texttt{p}(t')}, \quad \forall t' \in \mathcal{T}_B \backslash \{root\}; \\
    & a_{t'k}, d_{t'}  \in \{0,1\}, \quad \forall k \in [d], t' \in \mathcal{T}_B;
\end{align*}
\hrule
\vspace{0.3em}
\textbf{Fusion constraints for information sharing}:
\begin{align*}
    & (\gamma_{t_1j} - \gamma_{t_2j}) (1-r_{jt_1t_2})=0, \  \forall t_1, t_2 \in \mathcal{T}_L, \text{ and } t_1 \neq t_2; \\
    & r_{jt_1t_2} \in \{0,1\}, \  \forall j \in [2d+2],\  t_1, t_2 \in \mathcal{T}_L, \text{ and } t_1 \neq t_2.
\end{align*}
\end{algorithm}

\subsection{Causal analysis with fused optimal causal tree}\label{sec:sate esimtation}

A downstream goal in the subgroup analysis is to estimate the subgroup average treatment effect (SATE). Under standard causal assumptions
and model \eqref{regression}, the SATE for any subject $i$ can be given by
$$
\tau(\boldsymbol{X}_i,\Pi^*)=\sum_{m=1}^{M^*}\mathrm{1}_{(\boldsymbol{X}_i\in \mathcal{A}_m^*)}\cdot(\mu_m^*+\boldsymbol{\beta}^{*T}_{m} \bar{\boldsymbol{X}}_{m}).
$$
where $\bar{\boldsymbol{X}}_{m}=\mathbb{E}[\boldsymbol{X}_i|\boldsymbol{X}_i\in\mathcal{A}_m^*]$. The solution to optimization problem \eqref{optimization} yields a natural estimator of  $\tau(\boldsymbol{X}_i,\Pi^*)$:
$$
\hat{\tau}(\boldsymbol{X}_i,\hat{\Pi})=\sum_{m=1}^{{\hat{M}(\hat{\Pi})}}\mathrm{1}_{(\boldsymbol{X}_i\in \hat{\mathcal{A}}_m)}\cdot(\hat{\mu}_m+\hat{\boldsymbol{\beta}}_m^\top \hat{\bar{\boldsymbol{X}}}_m).
$$
where $\hat{\bar{\boldsymbol{X}}}_m$ is the sample average of $\boldsymbol{X}_i$ in $\hat{\mathcal{A}}_m$. To provide confidence intervals for $\tau(\boldsymbol{X}_i,\Pi^*)$ and enable valid statistical inference, we fix the subgroups and parameter fusion structure obtained from Algorithm \ref{algo:mio} and generate bootstrap samples $B$ times. For each bootstrap sample, we run the linear regression model with the parameter fusion pattern learned from the original sample, and then use the estimated coefficients and subgroup sample averages to estimate $\tau(\boldsymbol{X}_i,\Pi^*)$. Then we use the quantiles of these $B$ estimators to construct the confidence intervals.

To ensure good estimation performance of $\hat{\tau}(X_i,\hat{\Pi})$ on $\tau(X_i,\Pi^*)$, we need to ensure (1) subgroup identification accuracy: whether $\hat{\Pi}$ accurately recovers $\Pi^*$, (2) parameter estimation efficiency: whether $\hat{\boldsymbol{\gamma}}_m$ efficiently estimates $\boldsymbol{\gamma}_m$. As discussed in Section \ref{sec:contri}, our proposed fused optimal causal tree improves performance on both points compared to existing tree-based methods, leading to improved statistical efficiency. These advantages are demonstrated in the simulation results presented in Section \ref{sec:simulation}. In addition, a simple example is provided in Appendix \ref{app:eg fusion} to help illustrate this.

\section{THEORETICAL INVESTIGATION}\label{sec:theory}

In this section, we present a rigorous theoretical investigation of risk bounds for the optimal tree and greedy tree. We first introduce some notation. 
\subsection{Notations}
We consider the following non-parametric regression model:
$$
Y=f^*(\boldsymbol{Z})+\epsilon,
$$
where $\epsilon=Y-\mathbb{E}[Y|\boldsymbol{Z}]=Y-f^*(\boldsymbol{Z})$ is a sub-gaussian noise in the sense that there exists $\sigma^2>0$  such that for all $u\geq 0$,
$
\mathbb{P}(|\epsilon|\geq u) \leq \exp{(-u^2/(2\sigma^2))}.
$

In our regression model \eqref{regression}, we treat $\boldsymbol{Z}=[1,T,\boldsymbol{X}^\top,T\boldsymbol{X}^\top]^\top$, but the results in this section hold for more general forms of $\boldsymbol{Z} \in \mathbb{R}^p$. To evaluate the performance of a given tree regression approach, we use the following prediction risk measure:
\[
||f-f^*||_2^2 = \mathbb{E}_{\boldsymbol{Z}} \left[ \left(f^*(\boldsymbol{Z}) - f(\boldsymbol{Z})\right)^2 \right].
\]
We consider a function class of $h$-way interaction regression functions 
\begin{align*}
\mathcal{F}^{h}&:=\{f(\mathbf{Z})|f(\mathbf{Z}):=\sum_{j_{1}} f_{j_{1}}\left(Z_{j_{1}}\right)+ \cdots\\&+\sum_{j_{1}<j_{2}<\cdots<j_{h}} f_{j_{1}, j_{2}, ..., j_{h}}\left(Z_{j_1}, Z_{j_2},..., Z_{j_h}\right)\},
\end{align*}
and define the norm $||f||_\infty=\sup_{\boldsymbol{Z}\in \boldsymbol{\mathcal{Z}}}|f(\boldsymbol{Z})|$.

In addition, given a tree $$T_K=\{\{\hat{\mathcal{A}_j}\}_{j=1}^{M(T_K)},\{\hat{\boldsymbol{\gamma}}_j\}_{j=1}^{M(T_K)}\},$$ we denote the estimator of $f^*$ as
$$
\hat{f}(T_K)(\boldsymbol{Z})=\sum_{j=1}^{M(T_K)}\mathrm{1}_{(\boldsymbol{Z}\in \hat{\mathcal{A}}_j)}(\hat{\boldsymbol{\gamma}}_j)^T \boldsymbol{Z}.
$$
Where $\{\hat{\mathcal{A}_j}\}_{j=1}^{M(T_K)}$ is the collection of its leaf nodes and $\{\hat{\boldsymbol{\gamma}}_{j}\}_{j=1}^{M(T_K)} \in \mathbb{R}^p$ are the coefficient estimates obtained by fitting a linear model in each leaf node.
We also let \( \mathbb{E}_{\mathcal{D}_n}[\cdot] \) denote the expectation taken with respect to the training sample \( \mathcal{D}_n \).

\subsection{Theorectical results}
Now we present the out-of-sample risk bounds for both the optimal tree and greedy CART algorithm. 
\subsubsection{Risk bound for optimal tree}
\begin{assumption}\label{assum3}
Let $\boldsymbol{Z}\in [0,1]^p$ and $f^* \in \mathcal{F}^h$ is a piece-wise linear function induced by a depth $K$ tree $T_K^*=\{\{\mathcal{A}_j^*\}_{j=1}^{M(T_K^*)},\{\boldsymbol{\gamma}_j^*\}_{j=1}^{M(T_K^*)}\}$ in the sense that
$$
f^*(\boldsymbol{Z})=\sum_{j=1}^{M(T_K^*)}\mathrm{1}_{\{\boldsymbol{Z}\in \mathcal{A}_j^*\}}\boldsymbol{\gamma}^{*T}_j \boldsymbol{Z}.
 $$
 In addition, the $l_1$-norm of the coefficient vectors in each piece of $f^*$ are uniformly bounded by $B$ in the sense that 
$$
\|\boldsymbol{\gamma}_{j}\|_1=\sum_{k \in [p] }|\gamma_{jk}|\leq B,\quad \forall j=1,\ldots,M(T_K^*).
$$
\end{assumption}

\begin{theorem}[Risk bound for optimal tree]\label{thm:OT}
If $f^*$ satisfies Assumption \ref{assum3}, then under the sub-gaussian condition on $\epsilon$, we have
\begin{align*}
&\quad\mathbb{E}_{\mathcal{D}_n}\left(\left\|\hat{f}\left(T_{K}^{\mathrm{OT}}\right)-f^*\right\|_2^{2}\right)\\& \leq C_{1} \frac{2^{K} \log^2(N) \big((p+1)\log (N)+\log(p)\big)}{N}.
\end{align*}
where $T_{K}^{\mathrm{OT}}$ is a depth $K$ optimal tree obtained by Algorithm \ref{algo:mio}, $C_{1}$ is a positive constant depending only on B in Assumption \ref{assum3} and $\sigma^2$.
\end{theorem}
When the underlying true subgroup partition is known, the minimax risk lower bound for fitting a piecewise linear function with $2^K$ pieces is $\Omega(2^Kp\sigma^2/N)$. Theorem \ref{thm:OT} implies that, if the true underlying regression function is a piecewise linear function induced by a depth $K$ tree, then we can show the estimator based on the optimal regression tree approach can achieve near-optimal convergence rate with an additional logarithmic factor. The theorem is stated in terms of prediction risk, but it implicitly entails consistency of the estimated partition for subgroup structure. In particular, achieving the minimax rate in Theorem \ref{thm:OT} requires that the estimated partition asymptotically aligns (in measure) with the true partition. If an estimated leaf mixes two distinct true subgroups over a region with non-vanishing probability, the resulting approximation error would persist and prevent minimax-rate convergence.  An illustrated example is provided in Appendix \ref{example} to build intuition. 

\subsubsection{Risk bound for CART}
 To present the risk bound of CART, we need an additional technical assumption, which is used to handle the challenges in theoretical analysis caused by the greedy nature of CART. To save space, we provide this assumption and some comments in Appendix \ref{app:assumption}.
\begin{theorem}[Risk bound for CART]\label{thm:CART}
     Let $T_{K}^{\mathrm{CART}}$ be a depth $K$ tree obtained by CART with the splitting criteria in \eqref{criteria}. If $f^*$ and $T_K$ satisfy the Assumption \ref{assum3} and Assumption \ref{assum4}, then under the sub-gaussian condition on $\epsilon$, we have \begin{align*}
&\quad\mathbb{E}_{\mathcal{D}_n}\left(\left\|\hat{f}\left(T_{K}^{\mathrm{CART}}\right)-f^*\right\|_2^{2}\right) \leq \frac{C_0 h}{K+2h+1}+\\&C_{1} \frac{2^{K} \log^2(N) \big((p+1)\log (N)+\log(p)\big)}{N}.
\end{align*}
where $C_0$ is a positive constant depends on $f^*$, $C_{1}$ is a positive constant that depends only on $B$ in Assumption 1 and $\sigma^{2}$. 
\end{theorem}
Compared to the risk bound of the optimal tree, the risk bound of CART depends on an additional term $C_0\cdot h/
(K+2h+1)$. This means that CART needs depth $K \rightarrow \infty$ to consistently estimate the regression function $f^*$. Furthermore, this risk bound is derived under an additional stringent assumption (Assumption \ref{assum4}), which is often hard to verify in practice. The condition $K \to \infty$ required to ensure the consistency of CART inherently exacerbates overfitting, as increasing the tree depth introduces more parameters to estimate. In contrast, optimal regression trees do not rely on this condition for consistency, allowing for fewer parameters and mitigating overfitting. This further shows the superiority of optimal tree over greedy CART algorithm.

\section{SIMULATION} \label{sec:simulation}
\subsection{Simulation Setup}\label{sec:setup}
We consider the following data-generating process:
\begin{align*}
    & T\sim \text{Bernoulli}(p);\ X_1\sim \mathcal{N} (0,1); \ \boldsymbol{X}_{-1}\sim \mathcal{N} (\boldsymbol{0},\boldsymbol{\Sigma});
    \\& Y=\sum_{m=1}^{M^*} \mathrm{1}_{(\boldsymbol{X}\in \mathcal{A}_m^*)} (\delta_m+\mu_m T + \boldsymbol{\alpha}_m^{\top} \boldsymbol{X}+\boldsymbol{\beta}^\top_{m} T \boldsymbol{X})+\epsilon.
\end{align*}
where $\boldsymbol{X}_{-1}$ is the vector $\boldsymbol{X} \in \mathbb{R}^d$ without the first component. $\boldsymbol{\Sigma}$ is a matrix with $1$s on the diagonal and $\rho$ in the off-diagonal entries.  Here $\rho$ controls the correlation between some spurious split variables and the true split variable that determines the subgroup partitions. When $\rho$ is larger, it will be more difficult to identify the underlying subgroup partitions. Another two parameters $p$ and $d$ control the balance level of the treatment and control groups and the dimension of covariates $\boldsymbol{X}$. 
We test the performance of different methods for $\rho \in \{0.7,0.8\}$, $p=0.3$, and $d=3$ in the main manuscript and provide more simulations for other scenarios with different parameters in Appendix \ref{app:simu}. 

We set the number of subgroups $M^*=4$ such that $\mathcal{A}_1^*=\{\boldsymbol{X}|X_1<0,X_2<0\}$, $\mathcal{A}_2^*=\{\boldsymbol{X}|X_1<0,X_2\geq 0\}$,  $\mathcal{A}_3^*=\{\boldsymbol{X}|X_1\geq 0,X_2<0\}$ and $\mathcal{A}_4^*=\{\boldsymbol{X}|X_1\geq 0,X_2\geq 0\}$.
We provide the parameter specifications and the exact expressions of SATE, $\tau(\boldsymbol{x},\Pi^*,\rho)$, in Appendix \ref{append:sate parameter}.

\subsection{Methods}
\textbf{Causal Tree by CART (CT-CART):} We implement Breiman's original regression tree (CART) with a linear regression model in each leaf node and the splitting criteria discussed in \eqref{criteria} in the Appendix. 

\textbf{Optimal Causal Tree without Fusion (OCT):} The optimal causal tree is implemented by Algorithm \ref{algo:mio} without the fusion constraint. In other words, we set $\lambda=0$. We use Gurobi to solve this with $1$ hour limit. 

\textbf{Fused Optimal Causal Tree (FOCT):} The fused optimal causal tree is implemented by Algorithm \ref{algo:mio} with fusion constraints. We tune the $\lambda$ in the grid $
\Lambda=\{1/10000*i|i=1,\ldots,15\} 
$
and use BIC to choose the optimal $\lambda$. For each $\lambda$, we use Gurobi to solve this optimization problem with $1$ hour limit. 

For all these three methods, we set the tree depth as $2$ and the minimal leaf size as $N_{min}=1$. The collection of leaf nodes can serve as the identified subgroups and be used to estimate $\tau(\boldsymbol{x},\Pi^*,\rho)$ for any given observation $\boldsymbol{X}=\boldsymbol{x}$, which will be 
\begin{align*}
\hat{\tau}(\boldsymbol{x},\hat{\Pi}^{\text{me}},\mathcal{D}_n)=\sum_{m=1}^{\hat{M}^{\text{me}}}\mathrm{1}_{(\boldsymbol{x}\in \hat{\mathcal{A}}_m^{\text{me}})}\cdot(\hat{\mu}_m^{\text{me}}+(\hat{\boldsymbol{\beta}}_m^\text{me})^\top\hat{\bar{\boldsymbol{X}}}_m^{\text{me}}).
\end{align*}
where $\text{me} \in \{\text{CT-CART}, \text{OCT}, \text{FOCT}\}$ indicates the method we use.  $\hat{\Pi}^{\text{me}}=\{\hat{\mathcal{A}}_m^{\text{me}}\}_{m=1}^{\hat{M}^{\text{me}}}$ is the collection of leaf nodes for a given method. $\hat{\mu}_m^{\text{me}}$ and $\hat{\boldsymbol{\beta}}_m^\text{me}$ are the corresponding estimates in leaf node $\hat{\mathcal{A}}_m^{\text{me}}$ and $\hat{\bar{\boldsymbol{X}}}_m^{\text{me}}$ is the sample average of $\boldsymbol{X}$ in $\hat{\mathcal{A}}_m^{\text{me}}$. All of them are estimated by training data $\mathcal{D}_n$.

\subsection{Performance metrics}
We run 100 Monte Carlo simulations with each using  $n_{\text{train}}=200$ training samples from the data-generating process in Section \ref{sec:setup}, and compare three different performance metrics. 

\textbf{Subgroup identification accuracy:} We summarize the number of times that we identify the correct tree structures: depth 2 trees with the split variable at the root node is $X_1$ ($X_2$), followed by the split variable $X_2$ ($X_1$) in both its left and right child nodes.

\textbf{SATE estimation precision:} The precision of SATE \(\tau(\boldsymbol{x}, \Pi^*, \rho)\) estimates is evaluated by using the following mean squared error (MSE) metric:
\[
\frac{1}{n_{\text{test}}}\sum_{i=1}^{n_{\text{test}}}(\hat{\tau}(\boldsymbol{X}_i, \hat{\Pi}^{\text{me}}, \mathcal{D}_n) - \tau(\boldsymbol{X}_i, \Pi^*, \rho))^2.
\]
\textbf{Out-of-sample risk:} We also evaluate out-of-sample risk, which is the average of the squared differences between predicted values and actual values on test samples. We provide its detailed characterization in Appendix \ref{sec:out-sample-risk}.

For the latter two metrics, we use \(n_{\text{test}} = 2000 \) test samples generated from the same data-generating process to evaluate and compare the performance of different methods.
\begin{figure}[h]
    \centering
    \begin{subfigure}{0.22\textwidth}
      \includegraphics[width=\linewidth]{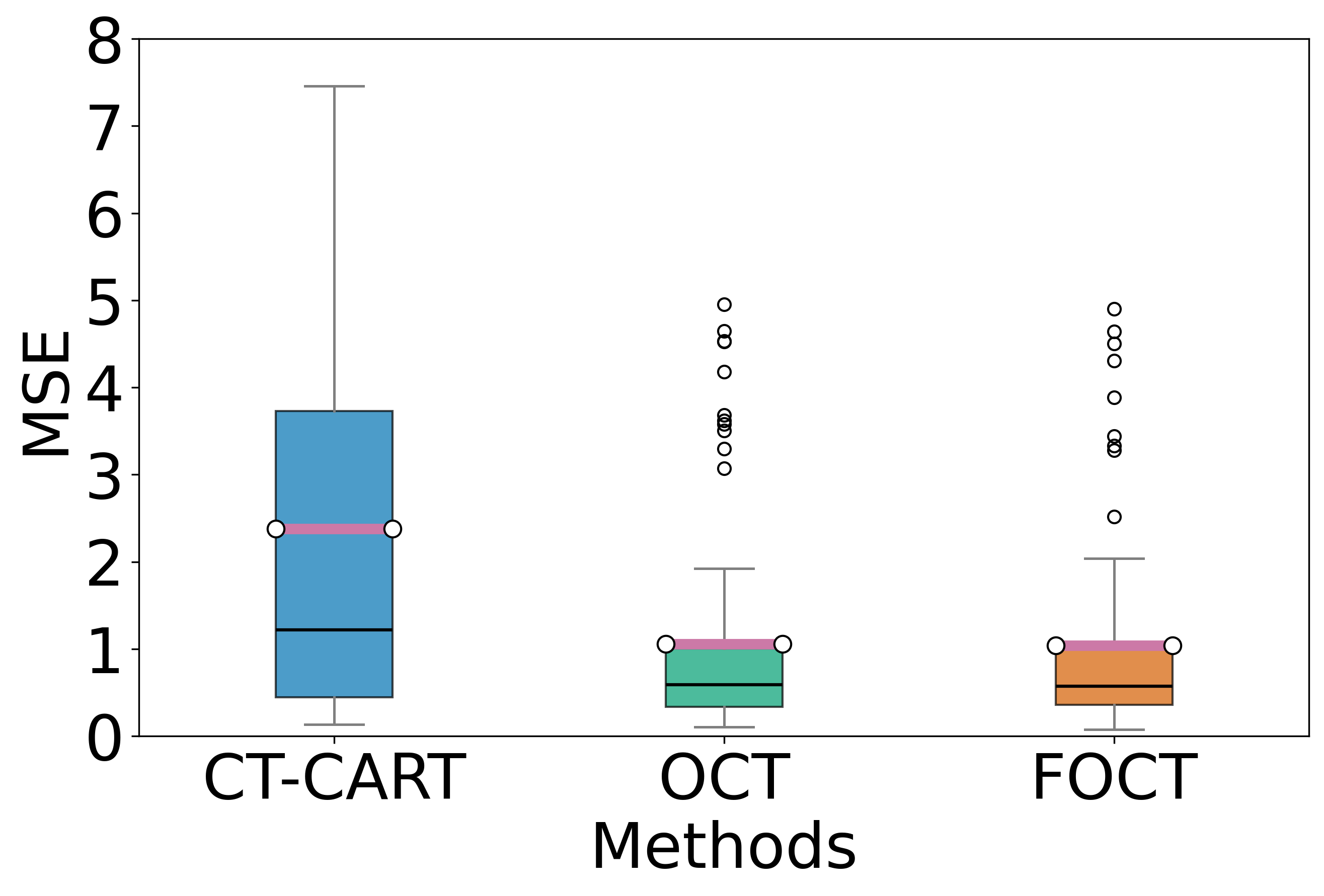}
        \caption{$\rho$=0.7}
        \label{fig:sate0.7}
    \end{subfigure}
    \hfill
    \begin{subfigure}{0.24\textwidth}       \includegraphics[width=\linewidth]{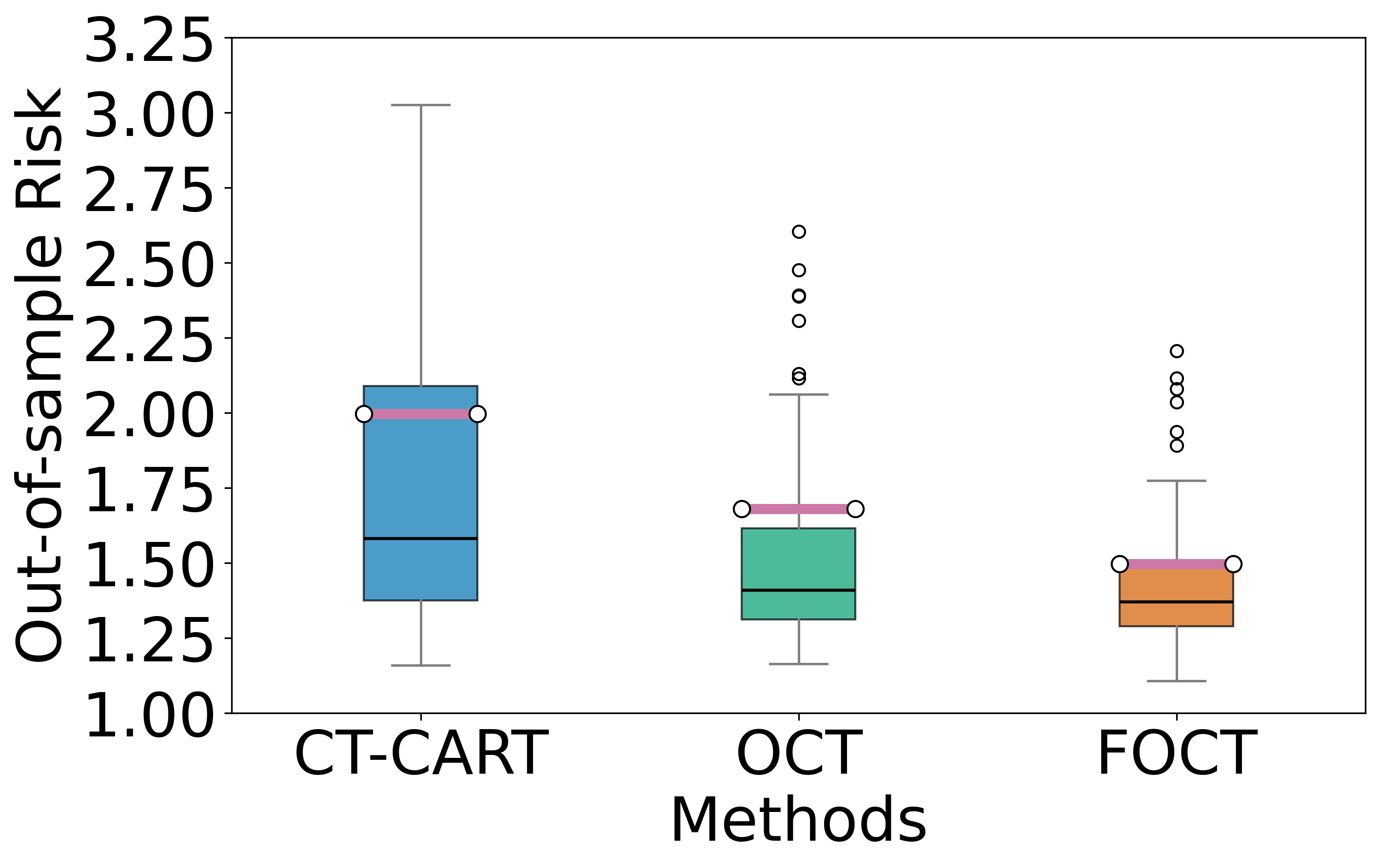}
        \caption{$\rho$=0.7}
        \label{fig:risk0.7}
    \end{subfigure}
    \centering
    \begin{subfigure}{0.23\textwidth}
      \includegraphics[width=\linewidth]{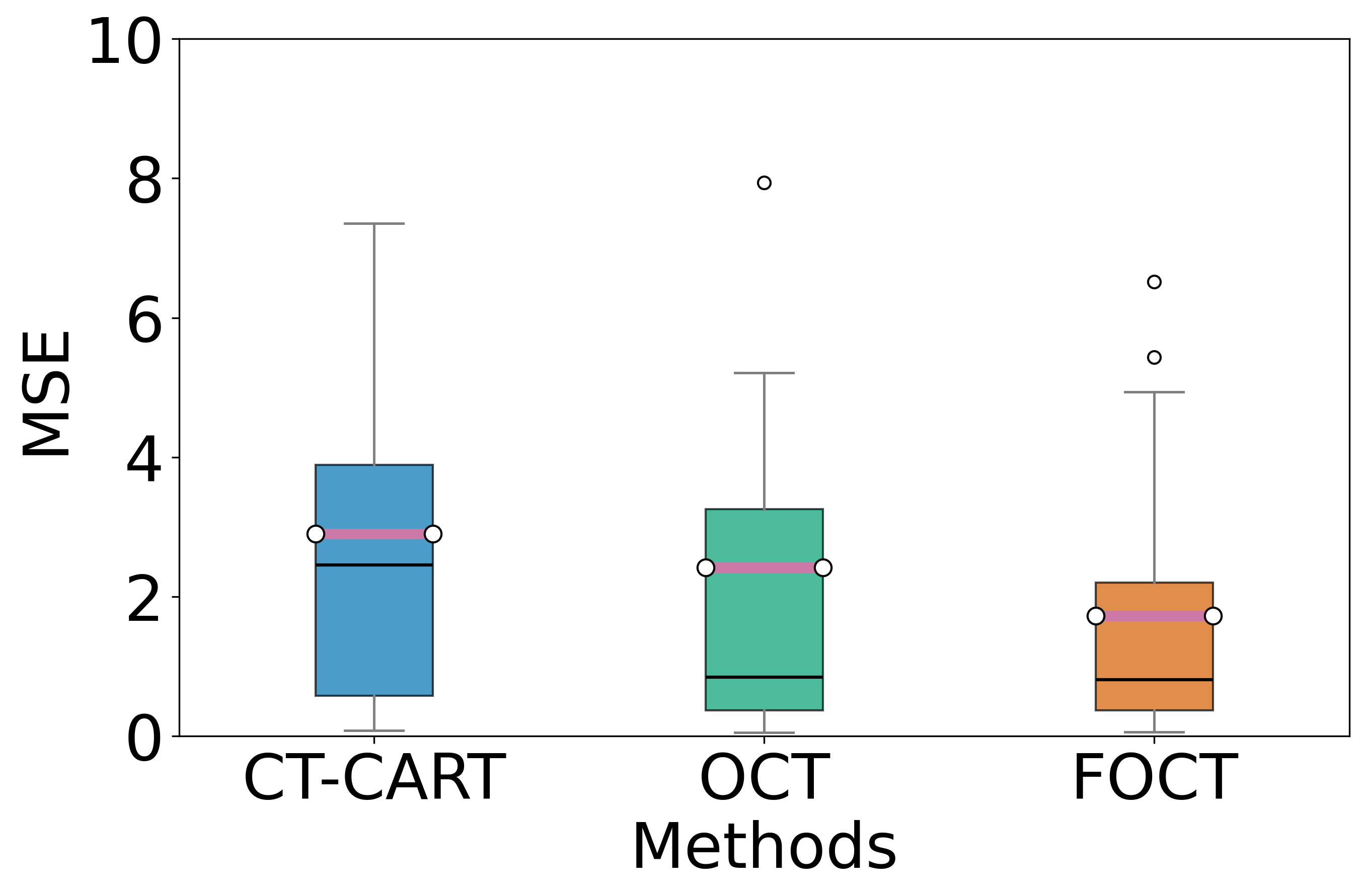}
        \caption{$\rho$=0.8}
        \label{fig:sate0.8}
    \end{subfigure}
    \hfill
    \begin{subfigure}{0.23\textwidth}       \includegraphics[width=\linewidth]{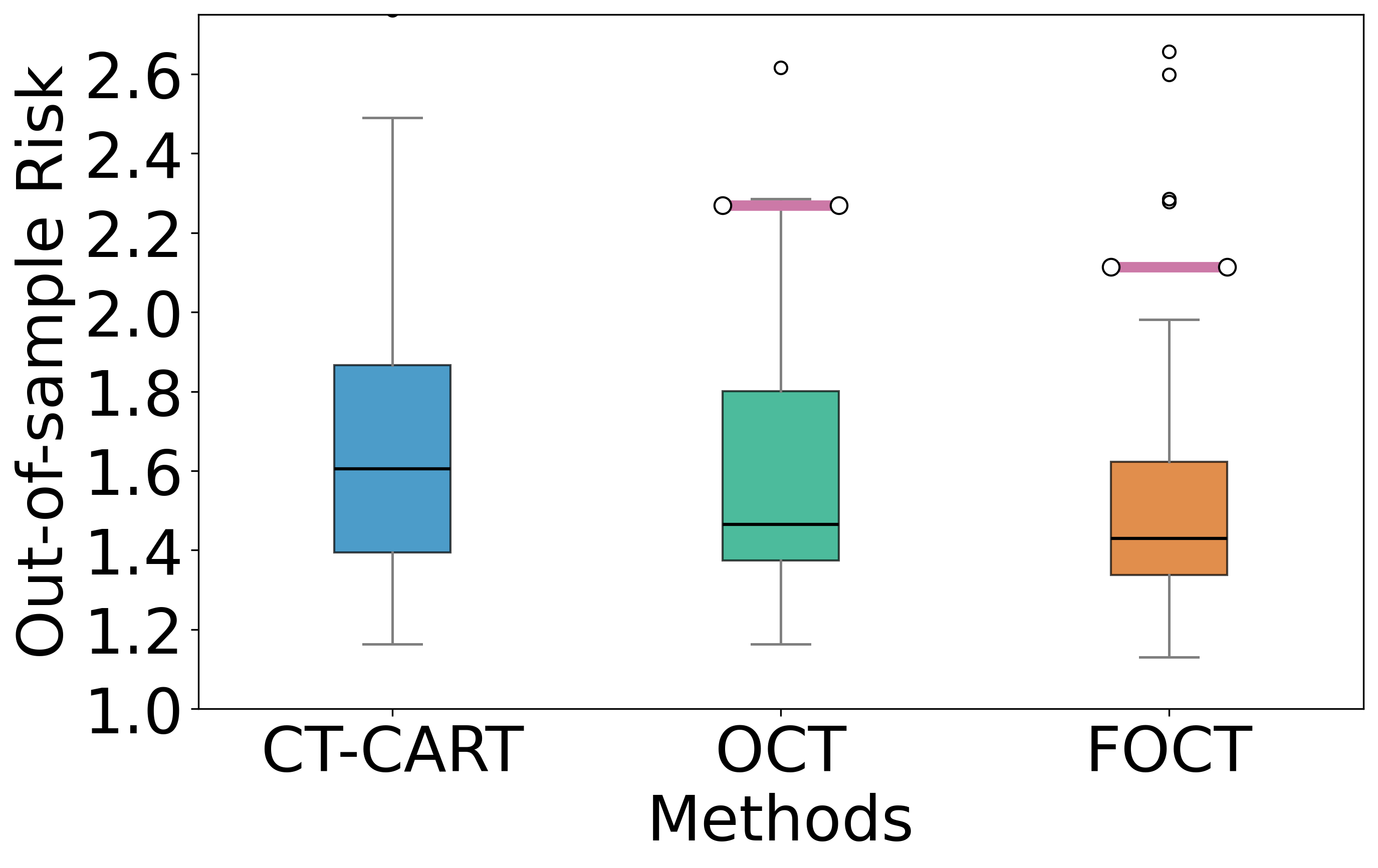}
        \caption{$\rho$=0.8}
        \label{fig:risk0.8}
    \end{subfigure}
    \caption{Boxplot of mean square error (MSE) of SATE estimation and out-of-sample risk for different methods with $\rho\in\{0.7,0.8\}$ across 100 Monte Carlo simulations. The purple line is the mean of MSE (or out-of-sample risk) across 100 Monte Carlo simulations for each method. The purple line for CT-CART in Figure \ref{fig:risk0.8} is above 2.7.}
\end{figure}
\subsection{Results}
When $\rho = 0.8$, FOCT more frequently recovers the correct tree structure (67 times) than OCT (54 times) and CT-CART (39 times), demonstrating the benefit of fusion constraints in improving subgroup partition accuracy by reducing overfitting and avoiding spurious splits like $X_3$. With a smaller $\rho = 0.7$, FOCT (81 times) and OCT (78 times) still outperform CT-CART (51 times), though the gap between FOCT and OCT narrows. This is expected, as stronger signals (smaller $\rho$) reduce overfitting, lessening the impact of the fusion constraint. Figures \ref{fig:sate0.7} and \ref{fig:risk0.7} show that when $\rho = 0.7$, both FOCT and OCT achieve lower MSE in SATE estimation and lower out-of-sample risk than CT-CART, with FOCT offering slightly more stable performance. When $\rho = 0.8$, identifying the true model becomes harder, and the fusion constraint plays a larger role in preventing overfitting and improving both subgroup recovery and estimation efficiency (Figures \ref{fig:sate0.8} and \ref{fig:risk0.8}).

\section{CASE STUDY}\label{sec:case study}

\textbf{Data:} We applied CT-CART, OCT, and FOCT to a dataset with black participants enrolled in the Health and Aging Brain Study–Health Disparities (HABS-HD). After excluding observations with missing data, the analytic sample comprised 237 individuals. The small size of this dataset underscores the necessity for statistically efficient subgroup discovery methods.  We chose a collection of covariates that potentially captures the AD heterogeneity (See Appendix \ref{sup:case} for details). For the treatment variable $T$, we assign individuals with $A\beta \geq 10$ to the group $T=1$. Those with $A\beta < 10$ are assigned to the group $T=0$. The outcome of interest, $Y$, is the Mini-Mental State Examination (MMSE) score, a widely used clinical measure of cognitive function for AD patients. Demographic characteristics of the analytic sample are summarized in Supplemental Table \ref{tab:demographics} in the Appendix. Implementation details are provided in the Appendix \ref{sup:case}. 

\textbf{Results:} The subgroups identified by FOCT are shown in Supplemental Figure \ref{fig:subgroups}, defined by neocortical tau levels ($Tau_2$), education (Edu), and age. As seen in Supplemental Figure \ref{fig:causal}, Group 3—individuals with low education ($Edu < 15$) and high neocortical tau ($Tau_2 \geq 1.071$)—shows a significant benefit from $A\beta$-lowering treatment, with improved MMSE scores (Negative values in Supplemental Figure \ref{fig:causal} reflect cognitive improvement as $A\beta$ levels decrease). This suggests that individuals in Group 3 may particularly benefit from $A\beta$-targeting therapies and should receive special attention in downstream AD trials. Supplemental Figure \ref{fig:coef} provides further insights into AD pathophysiology across subgroups. It reveals that age and education are strongly associated with MMSE: MMSE scores decline with age and increase with education. Notably, the positive effect of education on MMSE is more pronounced in Group 3, indicating a stronger protective role of cognitive reserve among individuals with low education and high tau burden. Additionally, tau accumulation in the neocortex ($Tau_2$) is significantly associated with MMSE both through its main effect and its interaction with treatment, with Group 3 showing the steepest decline in cognitive performance under high tau levels for $A\beta$-positive individuals. 

The subgroups identified by CT-CART and OCT are similar—but not identical—to the subgroups identified by FOCT (Supplemental Figure \ref{fig:subgroups_OCT}). However, without the parameter fusion constraint to mitigate overfitting, CT-CART and OCT fail to produce interpretable coefficient estimates for the effects of covariates on the outcome. In particular, for Group 2, both methods estimate a statistically significant positive effect of the interaction between treatment and APOE4 status on MMSE scores—contradicting well-established findings in the AD literature that APOE4 carriage is associated with cognitive decline (Supplemental Table \ref{tab:subgroup2}). Similarly, these two methods report a statistically significant negative effect of the interaction between treatment and APOE2, despite APOE2 being widely recognized as a protective allele in AD (Supplemental Table \ref{tab:subgroup2}). Furthermore, both CT-CART and OCT estimate a negative and statistically significant SATE of lowering A$\beta$
 on MMSE for Group 2. The bootstrap confidence intervals for SATEs under these methods are also substantially wider, reflecting a general loss of statistical efficiency compared to our FOCT approach (Supplemental Figure \ref{fig:causal_oct}). 

\section{DISCUSSION}\label{sec:discussion}
In this paper, we introduce a novel, efficient causal tree approach for subgroup analysis. Unlike existing tree-based methods, our approach identifies optimal subgroup partitions and enhances estimation power by incorporating a fusion constraint that accounts for the global structure of the data. We provide both theoretical analyses and empirical studies to demonstrate the superior performance of our method in terms of subgroup identification accuracy and causal effect estimation. We further illustrate its utility through a case study that yields interpretable clinical insights. Future directions include developing a more computationally efficient algorithm to alleviate the burden of solving the underlying mixed-integer optimization problem and extending the method to longitudinal data, which is essential for robust subgroup analysis \citep{wei2020precision}.

\bibliography{References}
\bibliographystyle{apalike}
\clearpage

\section*{Checklist}

\begin{enumerate}

  \item For all models and algorithms presented, check if you include:
  \begin{enumerate}
    \item A clear description of the mathematical setting, assumptions, algorithm, and/or model. [Yes]
    \item An analysis of the properties and complexity (time, space, sample size) of any algorithm. [Yes]
    \item (Optional) Anonymized source code, with specification of all dependencies, including external libraries. [Yes/No/Not Applicable]
  \end{enumerate}

  \item For any theoretical claim, check if you include:
  \begin{enumerate}
    \item Statements of the full set of assumptions of all theoretical results. [Yes]
    \item Complete proofs of all theoretical results. [Yes]
    \item Clear explanations of any assumptions. [Yes]     
  \end{enumerate}

  \item For all figures and tables that present empirical results, check if you include:
  \begin{enumerate}
    \item The code, data, and instructions needed to reproduce the main experimental results (either in the supplemental material or as a URL). [Yes]
    \item All the training details (e.g., data splits, hyperparameters, how they were chosen). [Yes]
    \item A clear definition of the specific measure or statistics and error bars (e.g., with respect to the random seed after running experiments multiple times). [Yes]
    \item A description of the computing infrastructure used. (e.g., type of GPUs, internal cluster, or cloud provider). [Yes]
  \end{enumerate}

  \item If you are using existing assets (e.g., code, data, models) or curating/releasing new assets, check if you include:
  \begin{enumerate}
    \item Citations of the creator If your work uses existing assets. [Yes]
    \item The license information of the assets, if applicable. [Not Applicable]
    \item New assets either in the supplemental material or as a URL, if applicable. [Not Applicable]
    \item Information about consent from data providers/curators. [Not Applicable]
    \item Discussion of sensible content if applicable, e.g., personally identifiable information or offensive content. [Not Applicable]
  \end{enumerate}

  \item If you used crowdsourcing or conducted research with human subjects, check if you include:
  \begin{enumerate}
    \item The full text of instructions given to participants and screenshots. [Not Applicable]
    \item Descriptions of potential participant risks, with links to Institutional Review Board (IRB) approvals if applicable. [Not Applicable]
    \item The estimated hourly wage paid to participants and the total amount spent on participant compensation. [Not Applicable]
  \end{enumerate}

\end{enumerate}

\clearpage
\appendix
\thispagestyle{empty}

\onecolumn
\section{Additional details of the case study}\label{sup:case}

\subsection{The chosen covariates} We chose covariates that potentially captures the AD heterogeneity, including, for example, sex, the apolipoprotein E2 (APOE2) allele, the apolipoprotein E4 (APOE4) allele, total years of education (Edu), age, tau accumulation in the entorhinal cortex ($Tau_1$), and tau accumulation in the neocortex ($Tau_2$). All included covariates have been linked to heterogeneity in AD pathologies: Female sex is associated with faster tau accumulation in vulnerable brain regions \citep{coughlan2025sex}.  APOE2 provides protection by reducing amyloid and tau burden \citep{kim2022apolipoprotein}, while APOE4 exacerbates pathology through increased amyloid-$\beta$ deposition, tau spread, and neuroinflammation \citep{fernandez2022apoe}. Higher education or cognitive reserve is linked to slower decline at equivalent pathology levels \citep{cody2024characterizing}.  Age is the strongest AD risk factor, with incidence rising sharply after 65 \citep{guerreiro2015age}. Tau accumulation in the entorhinal cortex predicts early memory loss \citep{cody2024characterizing}, and neocortical tau is a strong predictor of dementia progression \citep{sperling2024amyloid}.

\subsection{Implementation details}
We standardized all variables prior to analysis and applied our fused optimal causal tree with a maximum depth of 2 to the dataset. To select the tuning parameter for the fusion penalty, we generated a grid of 50 values evenly spaced over the interval $[0, 0.005]$. For each candidate value of $\lambda$, we used Gurobi to solve the optimization problem, imposing a one-hour time limit per run. The optimal $\lambda$ was selected based on BIC. In addition to identifying subgroups, the fused optimal causal tree reveals parameter fusion patterns, indicating which covariates exhibit homogeneous effects across some or all subgroups. Leveraging these identified fusion structures, we refitted a regression model with pooled coefficients and reported the estimated regression coefficients and their corresponding p-values in Table \ref{p-values}. We also reported the confidence intervals of coefficients in Figure \ref{fig:coef}. For SATE, we use the bootstrap method in Section \ref{sec:sate esimtation} with $B=1000$ to construct confidence intervals. We also compared the results of the other two methods, CT-CART and OCT, with our FOCT.

\newpage

\subsection{Additional figures and tables in the case study}

\begin{table}[ht]
\footnotesize
\centering
\begin{tabular}{lc} %
\hline
& \textbf{Black} \\ %
& N=237 (\%) \\ %
\hline
\textbf{Sex} & \\
Female & 145 (61.2) \\
Male & 92 (38.8) \\
\hline
\textbf{APOE2} & \\
0 & 195 (82.3) \\
1 & 40 (16.9) \\
2 & 2 (0.8) \\
\hline
\textbf{APOE4} & \\
0 & 154 (65.0) \\
1 & 73 (30.8) \\
2 & 10 (4.2) \\
\hline
\textbf{Education (years)} & \\
Min & 7 \\
Median & 14 \\
Max & 20 \\
\hline
\textbf{Age} & \\
50-55 & 34 (14.3) \\
55-60 & 55 (23.2) \\
60-65 & 59 (24.9) \\
65-70 & 47 (19.8) \\
70-75 & 32 (13.5) \\
75-80 & 5 (2.1) \\
80-85 & 3 (1.3) \\
85-90 & 2 (0.8) \\
\hline
\textbf{Amyloid-beta (Treatment)} & \\
$A\beta<10$ & 171 (72.2) \\
$A\beta\geq10$ & 66 (27.8) \\
\hline
\textbf{Tau in Medial Temporal Lobe ($Tau_1$)} & \\
Min & 0.675 \\
Median & 1.276 \\
Max & 2.215 \\
\hline
\textbf{Tau in the neocortex ($Tau_2$)} & \\
Min & 0.627 \\
Median & 1.066 \\
Max & 1.956 \\
\hline
\vspace{0.1cm}
\end{tabular}
\caption{Demographic characteristics of black participants in HABS-HD dataset}
\label{tab:demographics}
\end{table}

\begin{table}[ht]
\centering
\begin{tabular}{l|c|c}
\textbf{Variable} & \textbf{Estimate} & \textbf{P-Value} \\
\hline
Intercept & -0.031 & 0.813 \\
Treatment & 0.132 & 0.548 \\
Sex (group 4) & 0.195 & 0.445 \\
Sex (group 1/2/3) & -0.178 & 0.239 \\
APOE2 & -0.136 & 0.397 \\
APOE4  & 0.094 & 0.524 \\
Edu (group 3) & 1.298 & 0.000 \\
Edu (group 1/2/4) & 0.515 & 0.000 \\
Age  & -0.241 & 0.006 \\
$Tau_1$  & 0.025 & 0.842 \\
$Tau_2$  (group 1/3) & 0.590 & 0.011 \\
$Tau_2$ (group 2/4) & -0.328 & 0.108 \\
Treatment$\times$ Sex (group 2) & -0.214 & 0.582 \\
Treatment $\times$ Sex (group 1/3/4) & 0.360 & 0.287 \\
Treatment $\times$ APOE2 & -0.256 & 0.444 \\
Treatment $\times$ APOE4 & 0.238 & 0.386 \\
Treatment $\times$ Edu  & -0.177 & 0.356 \\
Treatment $\times$ Age & -0.018 & 0.903 \\
Treatment $\times$ $Tau_1$ (group 1/2) & 0.187 & 0.522 \\
Treatment $\times$ $Tau_1$ (group 3/4) & -0.292 & 0.191 \\
Treatment $\times$ $Tau_2$ (group 1/2/4) & 0.418 & 0.158 \\
Treatment $\times$ $Tau_2$ (group 3) & -0.962 & 0.000 \\
\hline
\vspace{0.1cm}
\end{tabular}
\caption{Coefficient Estimates and p-values for FOCT (Rounded to 3 Decimal Places)}
\label{p-values}
\end{table}

\begin{table}[ht]
\centering
\begin{tabular}{l|c|c}
\textbf{Variable} & \textbf{Estimate} & \textbf{P-value} \\
\hline
Intercept              &  0.368 & 0.555 \\
Treatment              &  0.576 & 0.809 \\
Sex                    & -0.367 & 0.219 \\
APOE2                     & -0.563 & 0.067 \\
APOE4               &  0.462 & 0.162 \\
Edu               &  0.525 & 0.018 \\
Age      &  0.163 & 0.340 \\
$Tau_1$    &  0.095 & 0.701 \\
$Tau_2$     &  0.126 & 0.865 \\
Treatment $\times$ Sex & -1.002 & 0.512 \\
Treatment $\times$ APOE2  & -1.443 & 0.310 \\
Treatment $\times$ APOE4 &  1.526 & 0.129 \\
Treatment $\times$ Edu & -0.401 & 0.518 \\
Treatment $\times$ Age &  0.999 & 0.254 \\
Treatment $\times$ $Tau_1$ &  0.414 & 0.679 \\
Treatment $\times$ $Tau_2$ &  0.333 & 0.725 \\
\hline
\end{tabular}
\caption{Coefficient estimates and p-values of subgroup 1 for CT-CART/OCT (Rounded to 3 Decimal Places)}
\label{tab:subgroup1}
\end{table}

\begin{table}[ht]
\centering
\begin{tabular}{l|c|c}
\textbf{Variable} & \textbf{Estimate} & \textbf{P-value} \\
\hline
Intercept              &  0.715 & 0.098 \\
Treatment              & -1.111 & 0.161 \\
Sex                    & -0.777 & 0.026 \\
APOE2                     &  0.628 & 0.146 \\
APOE4               &  0.113 & 0.712 \\
Edu               &  2.211 & 0.001  \\
Age      & -0.707 & 0.000  \\
$Tau_1$     &  0.323 & 0.274 \\
$Tau_2$    & -0.307 & 0.528 \\
Treatment $\times$ Sex &  1.345 & 0.011  \\
Treatment $\times$ APOE2  & -2.459 & 0.000  \\
Treatment $\times$ APOE4 &  1.347 & 0.020  \\
Treatment $\times$ Edu & -1.783 & 0.218 \\
Treatment $\times$ Age &  0.291 & 0.248 \\
Treatment $\times$ $Tau_1$  & -0.738 & 0.066 \\
Treatment $\times$ $Tau_2$  & -0.369 & 0.486 \\
\hline
\end{tabular}
\caption{Coefficient estimates and p-values of subgroup 2 for CT-CART/OCT (Rounded to 3 Decimal Places)}
\label{tab:subgroup2}
\end{table}

\begin{table}[ht]
\centering
\begin{tabular}{l|c|c}
\textbf{Variable} & \textbf{Estimate} & \textbf{P-value} \\
\hline
Intercept              &  0.016 & 0.916 \\
Treatment              &  0.467 & 0.158 \\
Sex                    &  0.247 & 0.151 \\
APOE2                     &  0.171 & 0.391 \\
APOE4               & -0.005 & 0.975 \\
Edu               &  0.393 & 0.010 \\
Age      & -0.072 & 0.448 \\
$Tau_1$      &  0.071 & 0.693 \\
$Tau_2$     & -0.220 & 0.294 \\
Treatment $\times$ Sex & -0.889 & 0.013 \\
Treatment $\times$ APOE2  & -0.158 & 0.719 \\
Treatment $\times$ APOE4 & -0.735 & 0.037 \\
Treatment $\times$ Edu & -0.141 & 0.639 \\
Treatment $\times$ Age &  0.063 & 0.753 \\
Treatment $\times$ $Tau_1$ &  0.081 & 0.777 \\
Treatment $\times$ $Tau_2$ & -0.348 & 0.193 \\
\hline
\end{tabular}
\caption{Coefficient estimates and p-values of subgroup 3 for CT-CART/OCT (Rounded to 3 Decimal Places)}
\label{tab:subgroup3}
\end{table}

\begin{figure}
    \centering
        \begin{subfigure}{\linewidth}
            \centering            \includegraphics[width=\linewidth]{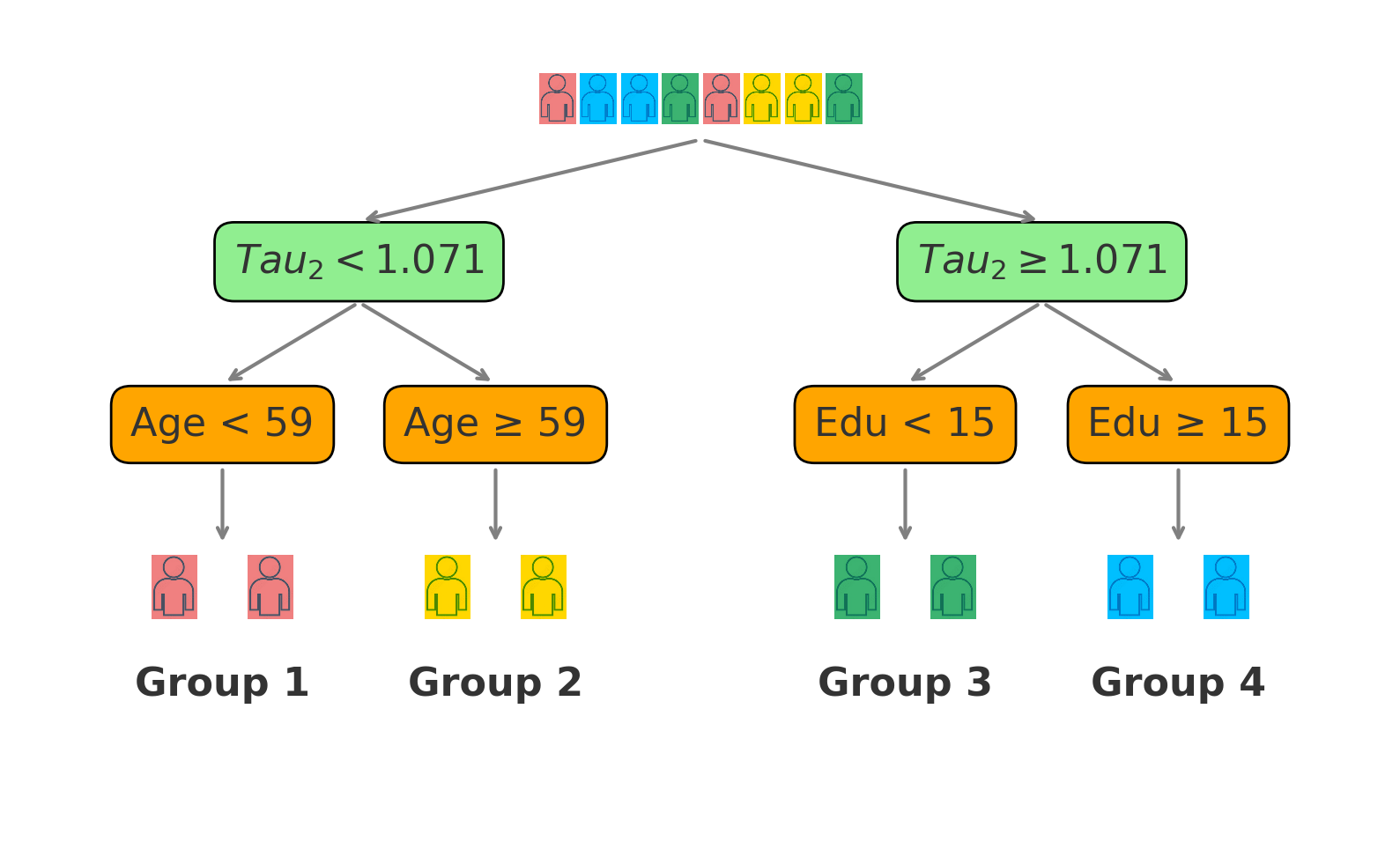}
            \caption{Subgroups identified by FOCT.}        \label{fig:subgroups}
        \end{subfigure}
        \begin{subfigure}{\linewidth}
            \centering            \includegraphics[width=0.85\linewidth]{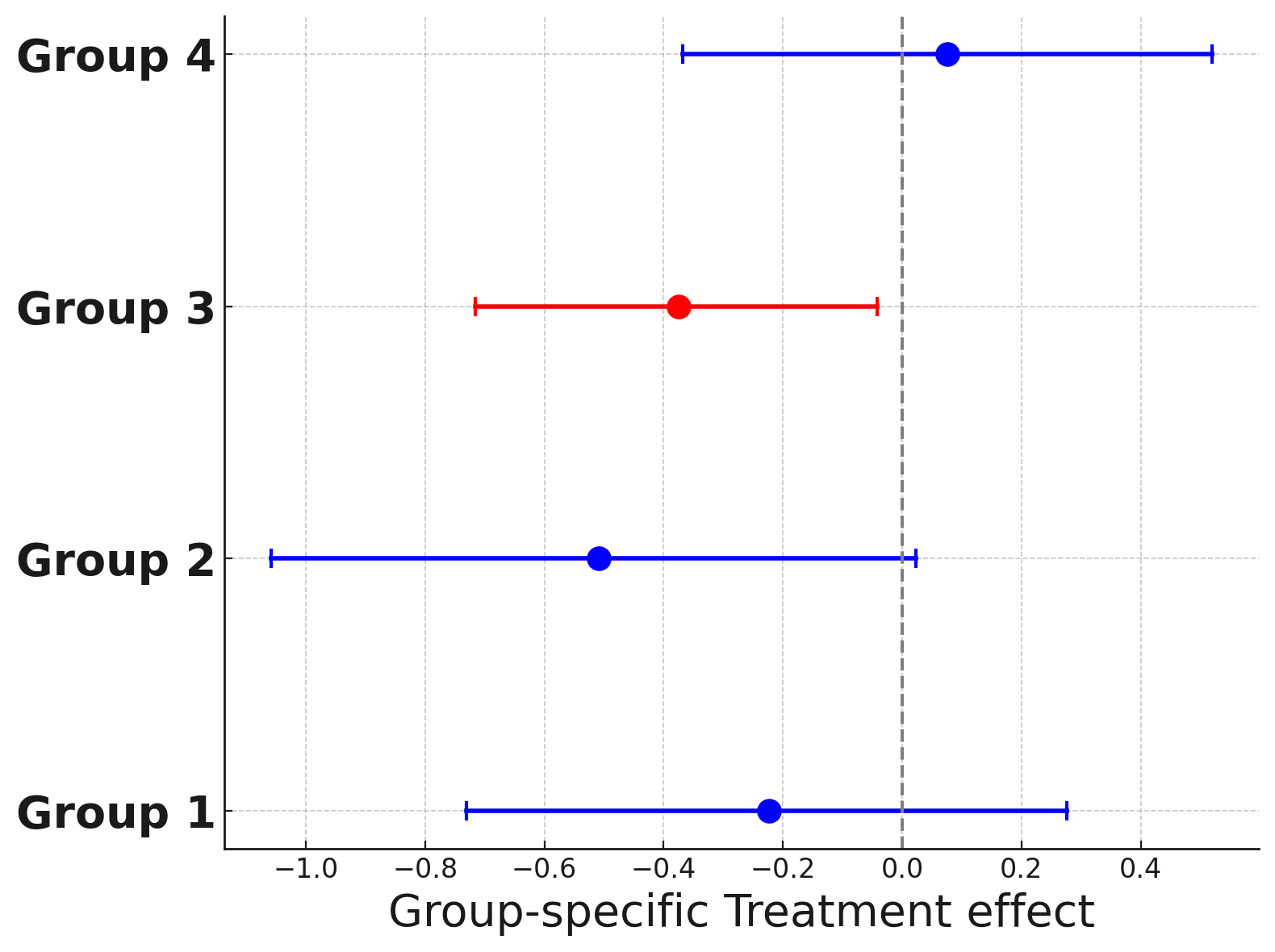}
            \caption{Group-specific treatment effects of $A\beta$ intervention on MMSE with 90\% confidence intervals. Red segment indicate significant effects. }        \label{fig:causal}
        \end{subfigure}
    \caption{Summary of findings by FOCT: HABS-HD case study on black individuals.}
\end{figure}

\begin{figure}
        \centering
            \vspace{0.3cm}
        \adjustbox{valign=c}{%
        \includegraphics[width=\linewidth]{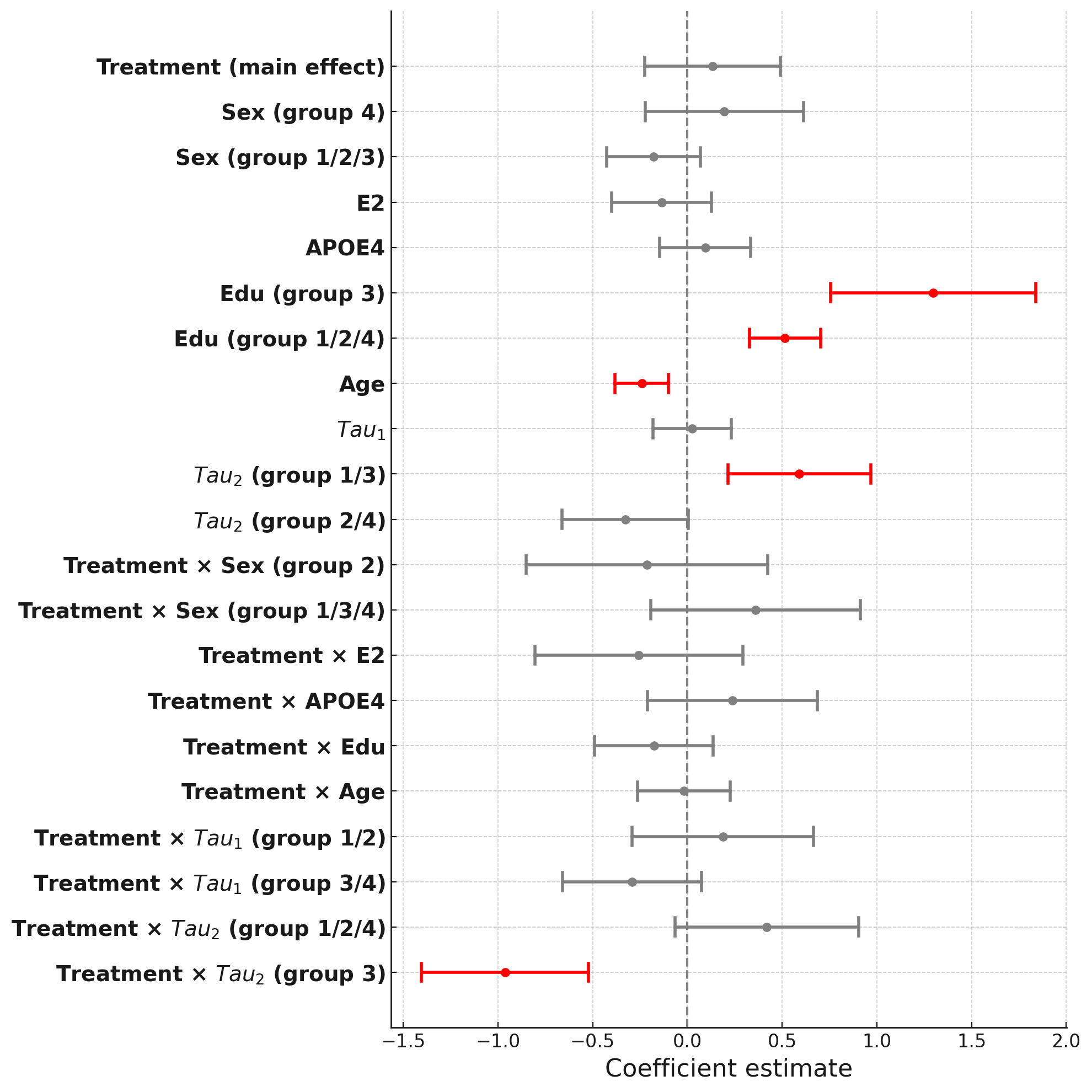}
        }
        \caption{Coefficient estimates in the regression model with parameter fusion pattern learned by FOCT with 90\% confidence intervals. Red segments indicate statistically significant coefficients; grey segments indicate non-significant ones.}\label{fig:coef}
\end{figure}

\begin{figure}
    \centering
        \begin{subfigure}{\linewidth}
            \centering            \includegraphics[width=\linewidth]{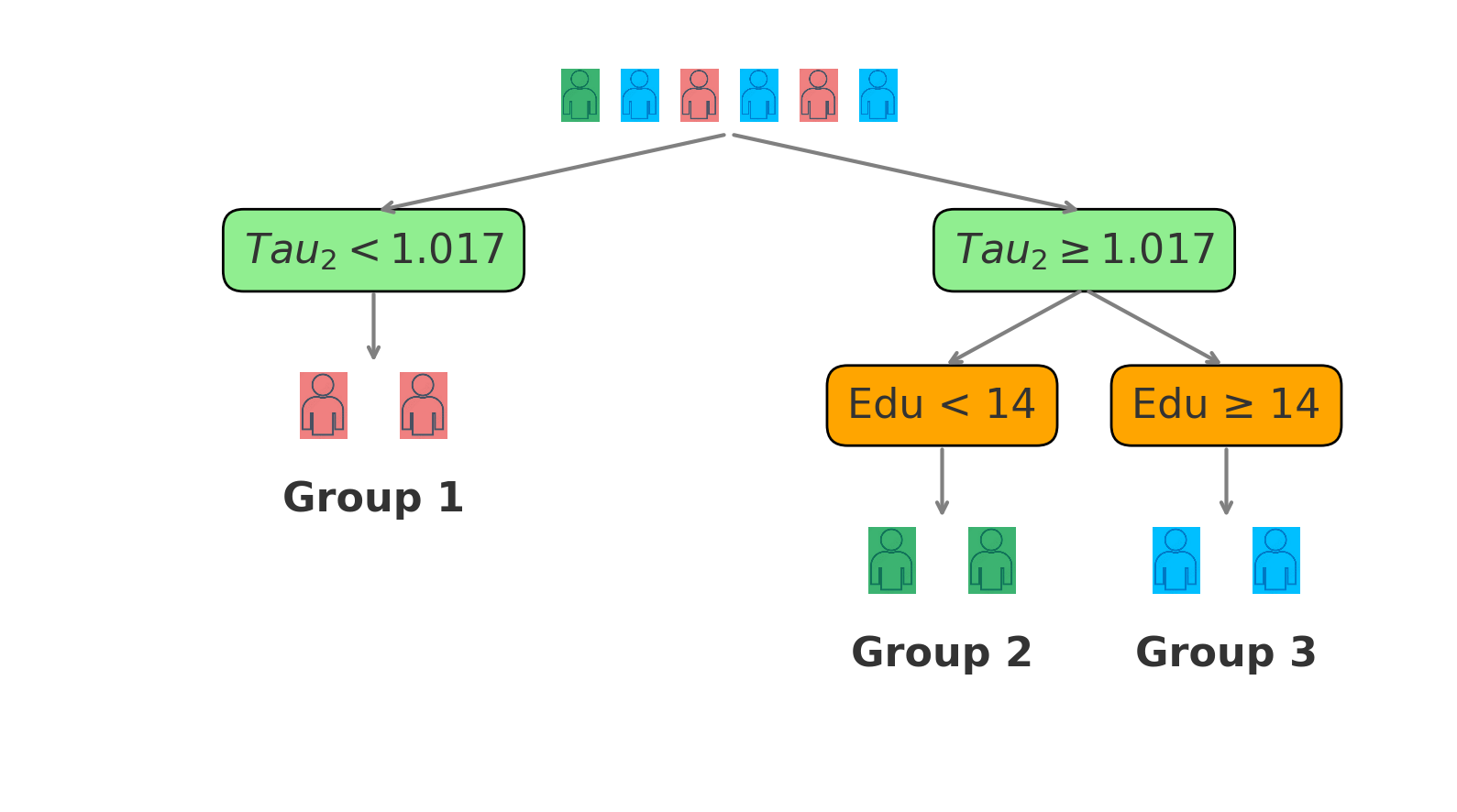}
            \caption{Subgroups identified by OCT/CT-CART.}        \label{fig:subgroups_OCT}
        \end{subfigure}
        \begin{subfigure}{\linewidth}
            \centering            \includegraphics[width=\linewidth,height=0.6\linewidth]{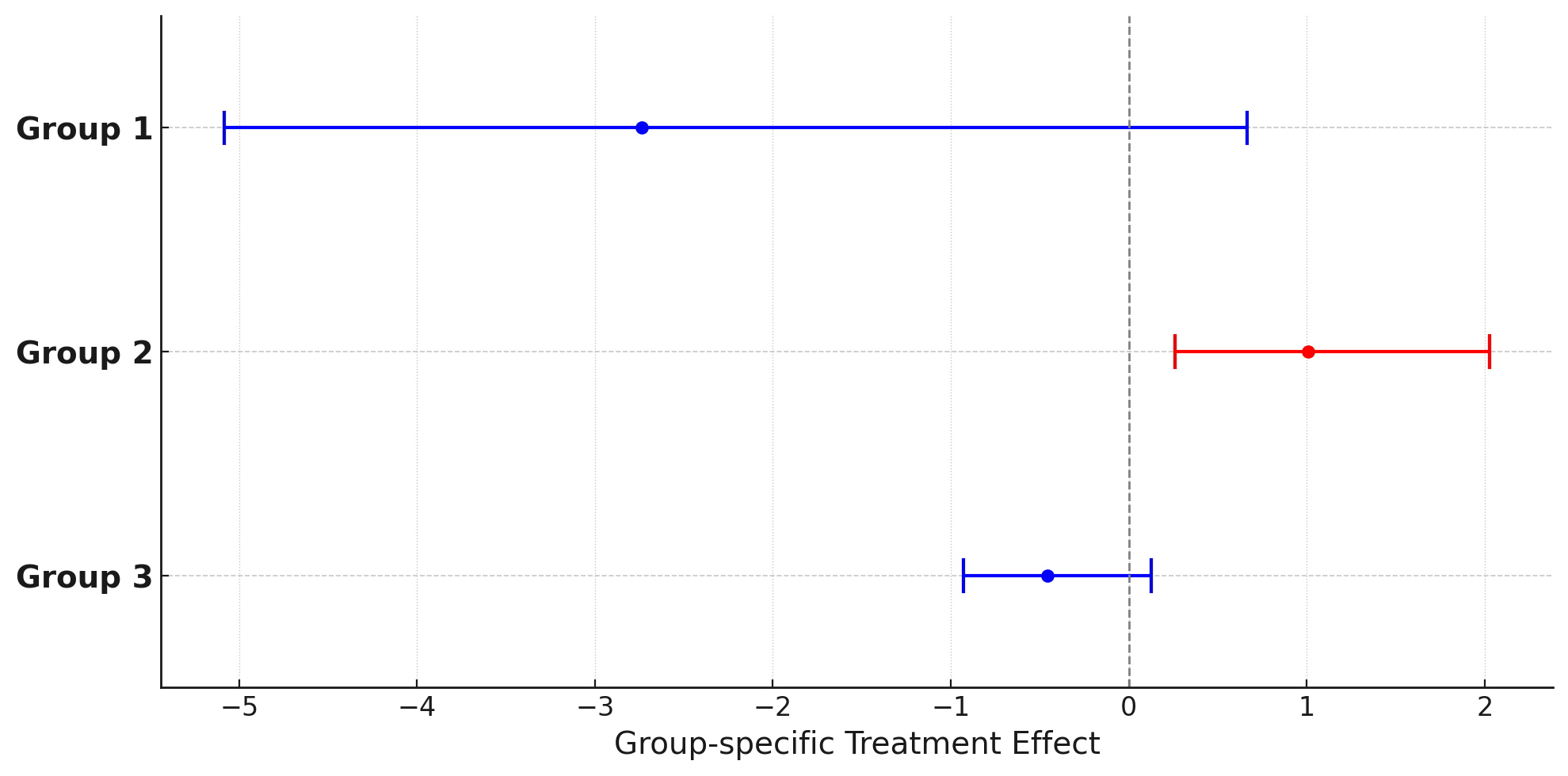}
            \caption{Group-specific treatment effects of $A\beta$ intervention on MMSE with 90\% confidence intervals. Red segment indicate significant effects. }        \label{fig:causal_oct}
        \end{subfigure}
    \caption{Summary of findings by OCT/CT-CART: HABS-HD case study on black individuals.}
\end{figure}
\clearpage

\section{Additional simulation details}\label{append:sate parameter}
\subsection{Parameters specification}
In our main simulation setup, we set the parameters as 

\begin{itemize}
    \item (1) $\delta_1=\delta_2=\delta_3=\delta_4=0$. 
    \item (2) $\mu_1=\mu_3=1$,  $\mu_2=\mu_4=2$.
    \item (3) $\boldsymbol{\alpha}_1=[1,1,2]^\top$, $\boldsymbol{\alpha}_2=[1,1,4]^\top$, $\boldsymbol{\alpha}_3=[1,1,3]^\top$, $\boldsymbol{\alpha}_4=[1,1,5]^\top$.
    \item (4) $\boldsymbol{\beta}_1=\boldsymbol{\beta}_2=[0,1,0]^\top$,\ $\boldsymbol{\beta}_3=\boldsymbol{\beta}_4=[1,1,2]^\top$.
\end{itemize}  

\subsection{Explicit forms of SATE}
The SATE for any given $\boldsymbol{X}_i$ is $\tau(\boldsymbol{X}_i,\Pi^*,\rho)=\sum_{m=1}^{4}\mathrm{1}_{(\boldsymbol{X}_i\in \mathcal{A}_m^*)}\cdot\Tilde{\mu}_m(\rho)$. Here $\Tilde{\mu}_1(\rho),\ldots,\Tilde{\mu}_4(\rho)$ are closed-form functions depending on $\rho$.
\begin{itemize}
    \item $\Tilde{\mu}_1(\rho)=1+\mathbb{E}[X_2|X_1<0,X_2<0]$,
    \item   $\Tilde{\mu}_2(\rho)=2+\mathbb{E}[X_2|X_1<0,X_2\geq0]$,
    \item   $\Tilde{\mu}_3(\rho)=1+\mathbb{E}[X_1+X_2+2X_3|X_1\geq 0,X_2<0]$,
    \item  $\Tilde{\mu}_4(\rho)=2+\mathbb{E}[X_1+X_2+2X_3|X_1\geq 0,X_2\geq0]$.
\end{itemize}
Here $X_1 \sim \mathcal{N}(0,1)$ and $(X_2,X_3)^T\sim \mathcal{N} (\boldsymbol{0},\boldsymbol{\Sigma})$ and $\boldsymbol{\Sigma}$ is a $2\times 2$  matrix with $1$s on the diagonal and $\rho$ in the off-diagonal entries. We have 
\begin{itemize}
    \item $\Tilde{\mu}_1(\rho)=1+\mathbb{E}[X_2|X_1<0,X_2<0]=1-\sqrt{\frac{2}{\pi}}$,
    \item   $\Tilde{\mu}_2(\rho)=2+\mathbb{E}[X_2|X_1<0,X_2\geq0]=2+\sqrt{\frac{2}{\pi}},$
            \item   $\Tilde{\mu}_3(\rho)=1+\mathbb{E}[X_1+X_2+2X_3|X_1\geq 0,X_2<0]=1- 2\rho\sqrt{\frac{2}{\pi}}$,
    \item  $\Tilde{\mu}_4(\rho)=2+\mathbb{E}[X_1+X_2+2X_3|X_1\geq 0,X_2\geq0]=2+ (2\rho+2)\sqrt{\frac{2}{\pi}}$.
\end{itemize}

\textbf{Calculation:}
Given \( X_2 \) is a standard normal random variable, the conditional expectation \( \mathbb{E}[X_2|X_1<0,X_2<0]=\mathbb{E}[X_2 | X_2 < 0] \) can be computed by using properties of the normal distribution. The expected value of a truncated normal distribution for a standard normal variable conditioned on \( X_2 < 0 \) is given by:
\[
\mathbb{E}[X_2 | X_2 < 0] = \frac{-\frac{1}{\sqrt{2\pi}}}{\frac{1}{2}} = -\sqrt{\frac{2}{\pi}}.
\]
Similarly 
\[
\mathbb{E}[X_2 | X_2 \geq 0] = \frac{-\frac{1}{\sqrt{2\pi}}}{\frac{1}{2}} = \sqrt{\frac{2}{\pi}}.
\]
Then we have
\begin{itemize}
    \item $\Tilde{\mu}_1(\rho)=1+\mathbb{E}[X_2|X_1<0,X_2<0]=1-\sqrt{\frac{2}{\pi}}$,
    \item   $\Tilde{\mu}_2(\rho)=2+\mathbb{E}[X_2|X_1<0,X_2\geq0]=2+\sqrt{\frac{2}{\pi}}$.
\end{itemize}
\begin{itemize}
\item   $\Tilde{\mu}_3(\rho)=1+\mathbb{E}[X_1+X_2+2X_3|X_1\geq 0,X_2<0]$,
    \item  $\Tilde{\mu}_4(\rho)=2+\mathbb{E}[X_1+X_2+2X_3|X_1\geq 0,X_2\geq0]$.
\end{itemize}
we first calculate
$$\mathbb{E}[X_1+X_2+2X_3|X_1\geq 0,X_2<0]=\mathbb{E}[X_2+2X_3|X_2<0]+\mathbb{E}[X_1|X_1\geq 0].$$
To calculate the conditional expectation:
\[
\mathbb{E}[X_2 + 2X_3 \mid X_2 < 0],
\]
where \( X_2 \) and \( X_3 \) are jointly standard normal random variables with correlation coefficient \( \rho \), we decompose the expectation into the sum of individual expectations:
\[
\mathbb{E}[X_2 + 2X_3 \mid X_2 < 0] = \mathbb{E}[X_2 \mid X_2 < 0] + 2\mathbb{E}[X_3 \mid X_2 < 0].
\]
Since \( X_2 \) and \( X_3 \) are jointly normal, the conditional distribution of \( X_3 \) given \( X_2 \) is normal with \( \mathbb{E}[X_3 \mid X_2] = \rho X_2 \).
To find \( \mathbb{E}[X_3 \mid X_2 < 0] \), we take the expectation over all \( X_2 < 0 \):

\[
\mathbb{E}[X_3 \mid X_2 < 0] = \mathbb{E}_{X_2}\left[ \mathbb{E}[X_3 \mid X_2] \mid X_2 < 0 \right] = \mathbb{E}_{X_2}\left[ \rho X_2 \mid X_2 < 0 \right]=\rho \mathbb{E}[X_2 \mid X_2 < 0].
\]
Now, we substitute the computed expectations back into the original expression:
\begin{align*}
\mathbb{E}[X_2 + 2X_3 \mid X_2 < 0] &= \mathbb{E}[X_2 \mid X_2 < 0] + 2\mathbb{E}[X_3 \mid X_2 < 0] \\
&= (1+2\rho)\cdot\mathbb{E}[X_2 \mid X_2 < 0]\\
&= - \sqrt{\frac{2}{\pi}} (1 + 2\rho).
\end{align*}
With $\mathbb{E}[X_1 \mid X_1 \geq 0] =  \sqrt{\frac{2}{\pi}}$ we have 
$$
\mathbb{E}[X_1+X_2+2X_3|X_1\geq 0,X_2<0]=- 2\rho\sqrt{\frac{2}{\pi}} . 
$$
Similarly, 
$$
\mathbb{E}[X_1+X_2+2X_3|X_1\geq 0,X_2\geq0]= (2\rho+2)\sqrt{\frac{2}{\pi}}.
$$
Thus we can show
\begin{itemize}
        \item   $\Tilde{\mu}_3(\rho)=1+\mathbb{E}[X_1+X_2+2X_3|X_1\geq 0,X_2<0]=1- 2\rho\sqrt{\frac{2}{\pi}}$,
    \item  $\Tilde{\mu}_4(\rho)=2+\mathbb{E}[X_1+X_2+2X_3|X_1\geq 0,X_2\geq0]=2+ (2\rho+2)\sqrt{\frac{2}{\pi}}$.
\end{itemize}

\subsection{Out-of-sample risk evaluation} \label{sec:out-sample-risk}
The out-of-sample risk is evaluated on the test sample by
\[
\frac{1}{n_{\text{test}}}\sum_{i=1}^{n_{\text{test}}}(\hat{f}(\boldsymbol{Z}_i, \hat{T}^{\text{me}}, \mathcal{D}_n) - f^*(\boldsymbol{Z}_i))^2.
\]
Here $\hat{T}^{\text{me}}$ is the tree found by method $\text{me} \in \{\text{CT-CART}, \text{OCT}, \text{FOCT}\}$. In addition, $\hat{f}(\boldsymbol{Z}_i, \hat{T}^{\text{me}}, \mathcal{D}_n)$ is the estimated model with tree $\hat{T}^{\text{me}}$ and $f^*(\boldsymbol{Z}_i)$ is the true mean under model \eqref{regression}.

\newpage
\section{Additional Experiments}\label{app:simu}

\subsection{Smaller sample size $n=100$ (OCT can perform poorly in extremely small samples).}

We also conduct a series of simulations with smaller sample sizes, where overfitting issues become more pronounced. Under these conditions, the advantages of adopting FOCT are further underscored. We tune the $\lambda$ in the grid $
\Lambda=\{1/500*i|i=1,\ldots,5\}  
$ in this section. 

\subsubsection{The dimension of variables $d=3$, imbalanced treatment and control groups with $p=0.3$, and $\rho=0.5$}\label{n=100,d=3}

\begin{figure}[h]
    \centering
    \begin{subfigure}{0.48\textwidth}
      \includegraphics[width=\linewidth]{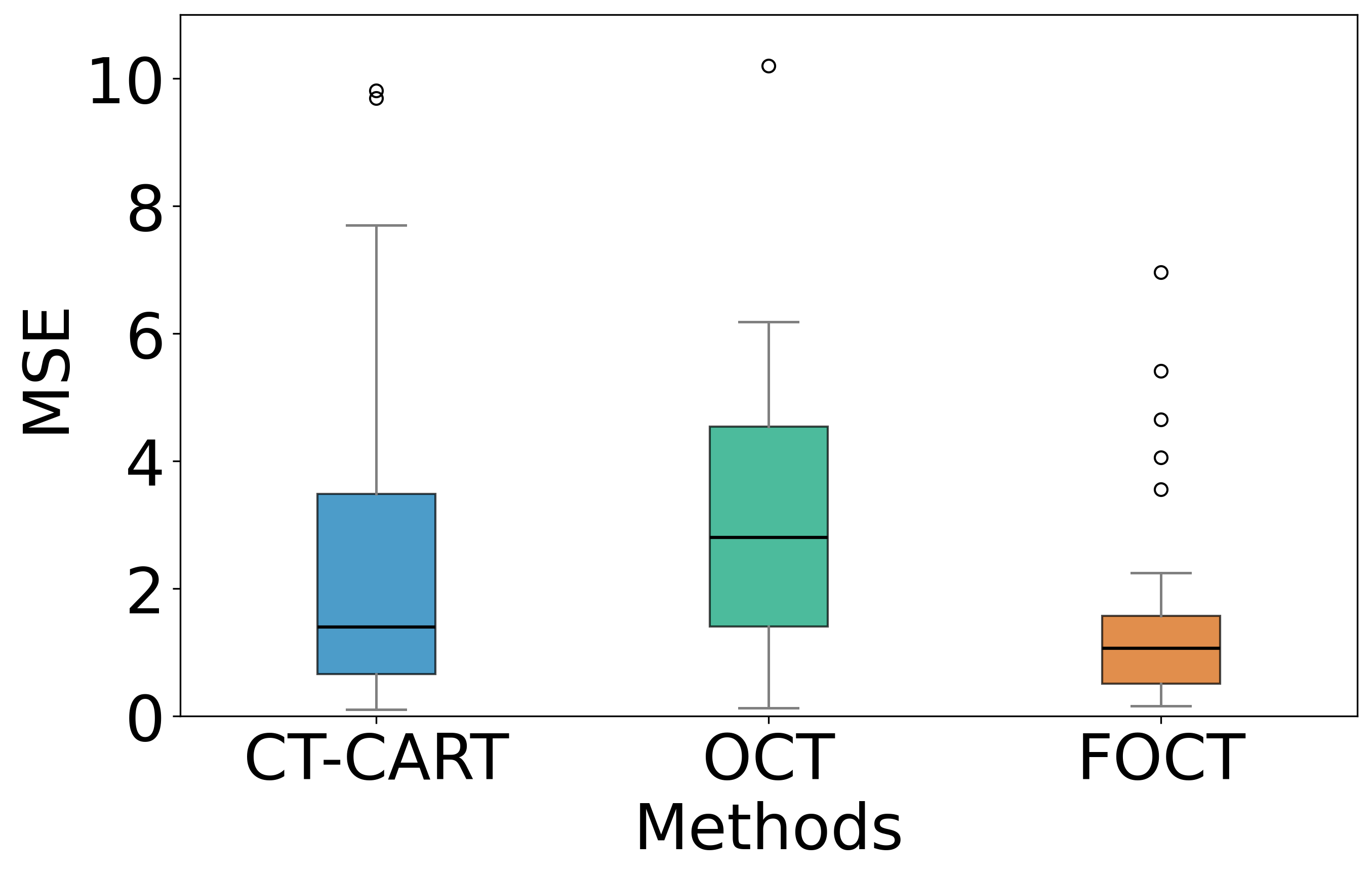}
        \caption{$n=100$, $p=0.3$, $\rho=0.5$ and $d=3$}
        \label{fig:saten100d3}
    \end{subfigure}
    \hfill
    \begin{subfigure}{0.48\textwidth}       \includegraphics[width=\linewidth]{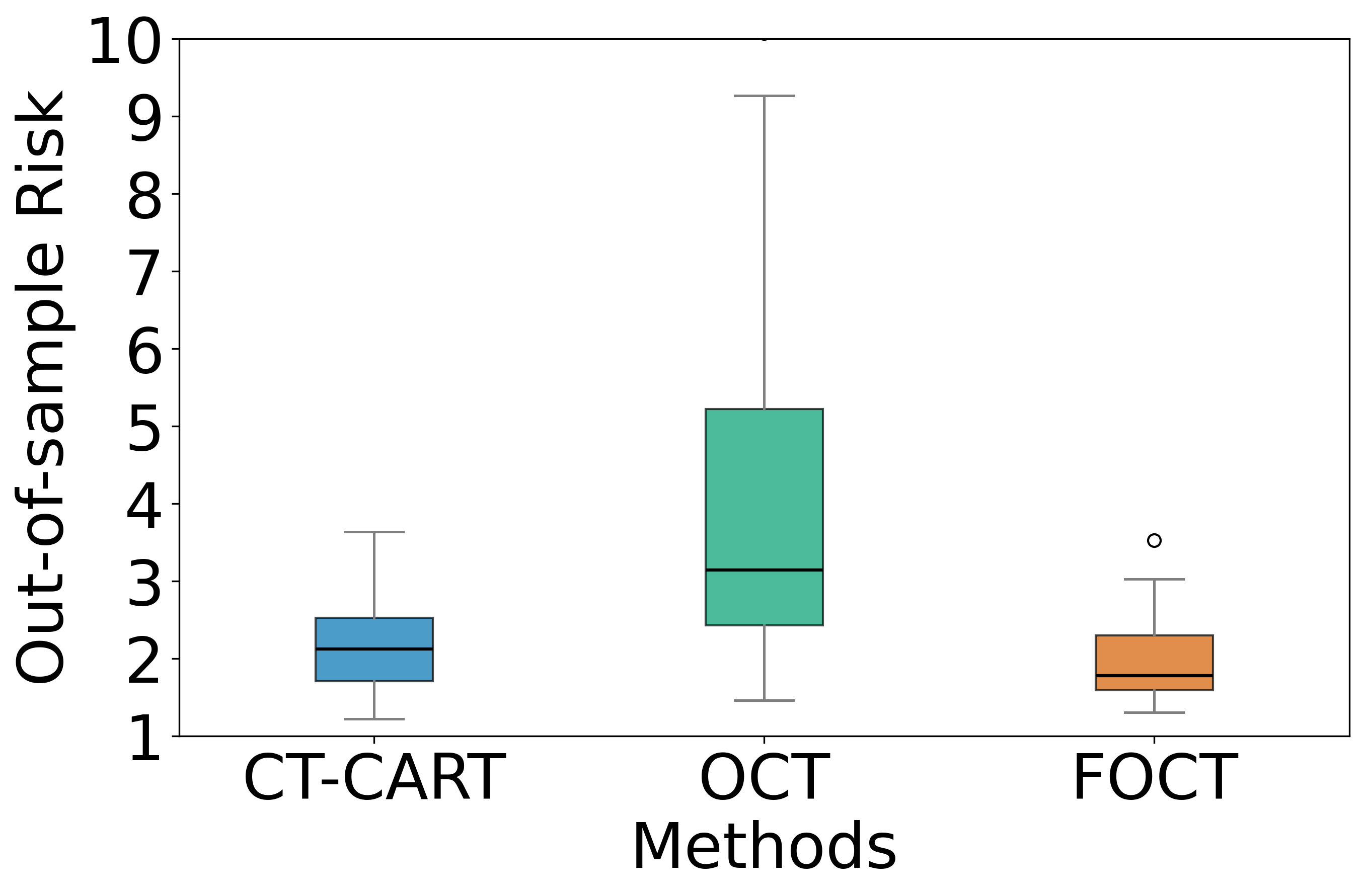}
        \caption{$n=100$, $p=0.3$, $\rho=0.5$ and $d=3$}
        \label{fig:riskn100d3}
    \end{subfigure}
    \vspace{0.3in}
    \caption{Boxplot of mean square error (MSE) of SATE estimation and out-of-sample risk for different methods with $n=100$, $p=0.3$, $\rho=0.5$ and $d=3$ across 50 Monte Carlo simulations.}
\end{figure}

We reduce the sample size to \(n=100\), thereby exacerbating the overfitting issue. Across 50 Monte Carlo simulations, CT-CART identifies the correct tree structure only 10 times, OCT 20 times, and FOCT 39 times—indicating a clear advantage for FOCT. In terms of SATE estimation precision and out-of-sample risk, OCT performs the worst, while FOCT performs the best (Figure \ref{fig:saten100d3} and Figure \ref{fig:riskn100d3}). This disparity arises because, with such a small sample size, overfitting is extreme: OCT, which optimizes the tree structure solely based on training data, suffers most from overfitting, resulting in poor generalization. In contrast, the fusion constraint in FOCT mitigates overfitting and improves out-of-sample performance. Although CT-CART is less prone to overfitting in this scenario, it struggles to accurately identify subgroups. Overall, FOCT outperforms the other methods in this scenario, achieving higher subgroup identification accuracy, better SATE estimation, and lower out-of-sample risk.

\subsubsection{The dimension of variables $d=5$, imbalanced treatment and control groups with $p=0.3$, and $\rho=0.5$}

\begin{figure}[h]
    \centering
    \begin{subfigure}{0.48\textwidth}
      \includegraphics[width=\linewidth]{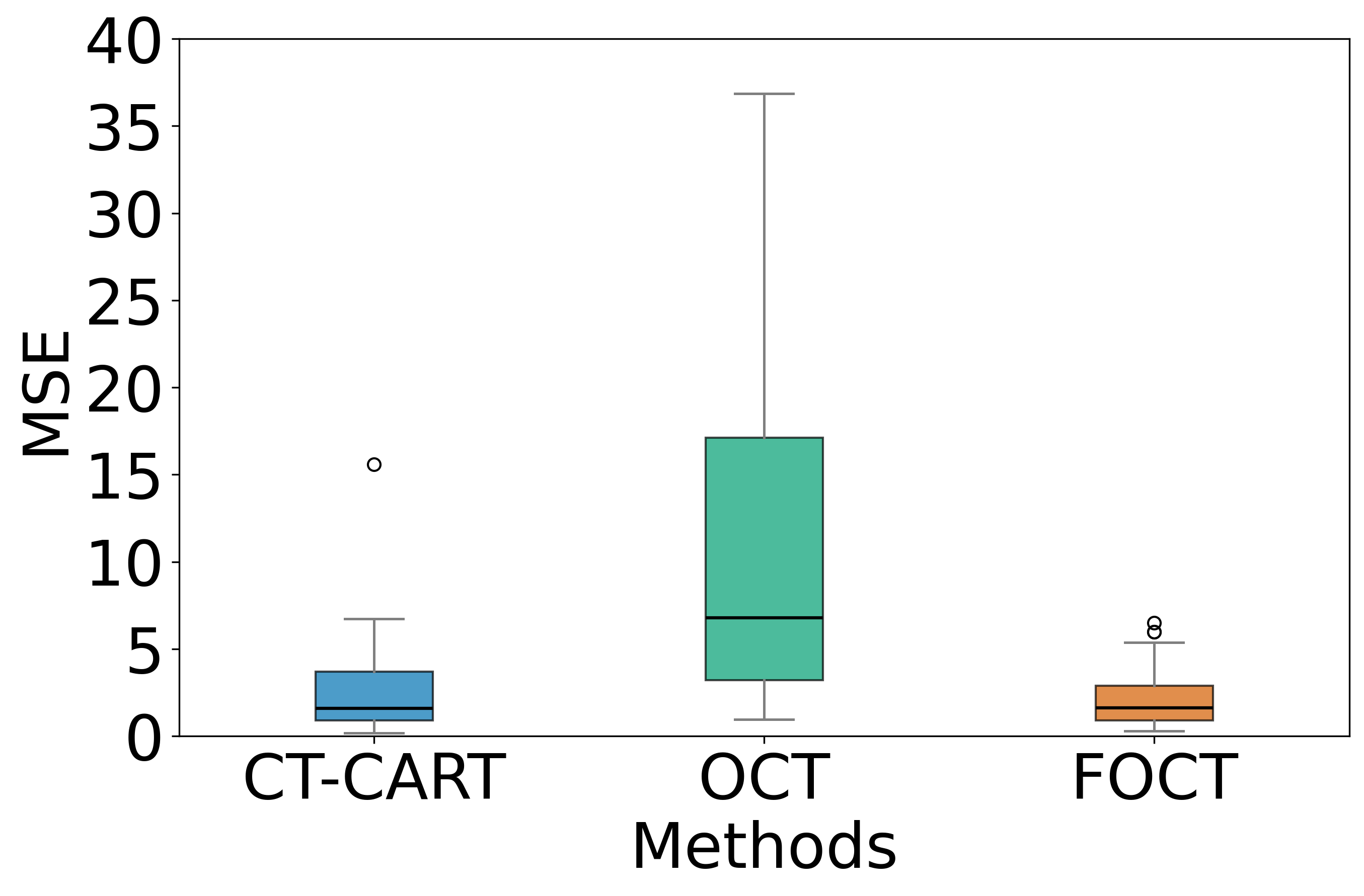}
        \caption{$n=100$, $p=0.3$, $\rho=0.5$ and $d=5$}
        \label{fig:saten100d5}
    \end{subfigure}
    \hfill
    \begin{subfigure}{0.48\textwidth}       \includegraphics[width=\linewidth]{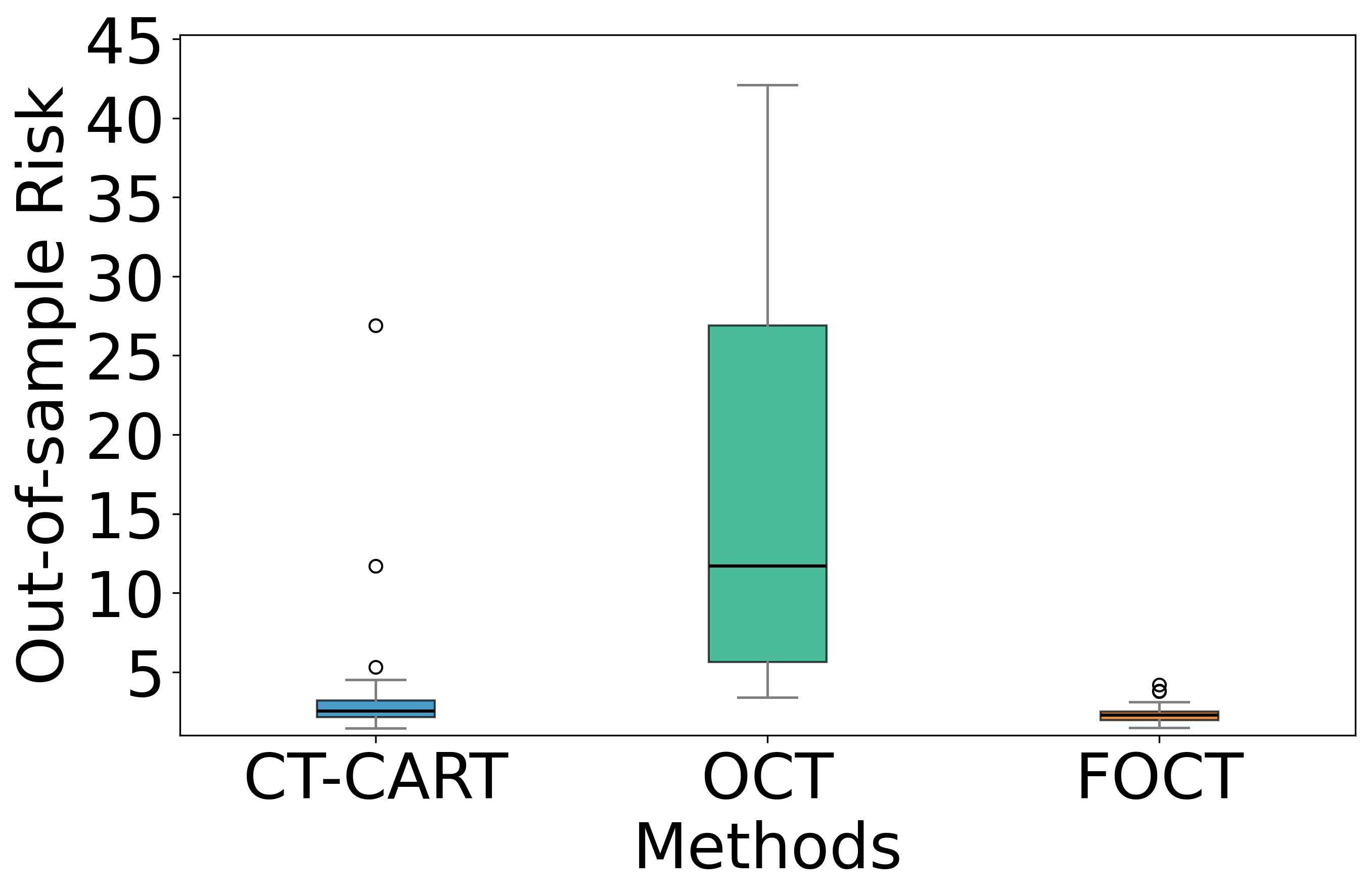}
        \caption{$n=100$, $p=0.3$, $\rho=0.5$ and $d=5$}
        \label{fig:riskn100d5}
    \end{subfigure}
    \vspace{0.3in}
    \caption{Boxplot of mean square error (MSE) of SATE estimation and out-of-sample risk for different methods with $n=100$, $p=0.3$, $\rho=0.5$ and $d=5$ across 50 Monte Carlo simulations.}
\end{figure}

\begin{figure}[h]
    \centering
    \begin{subfigure}{0.48\textwidth}
      \includegraphics[width=\linewidth]{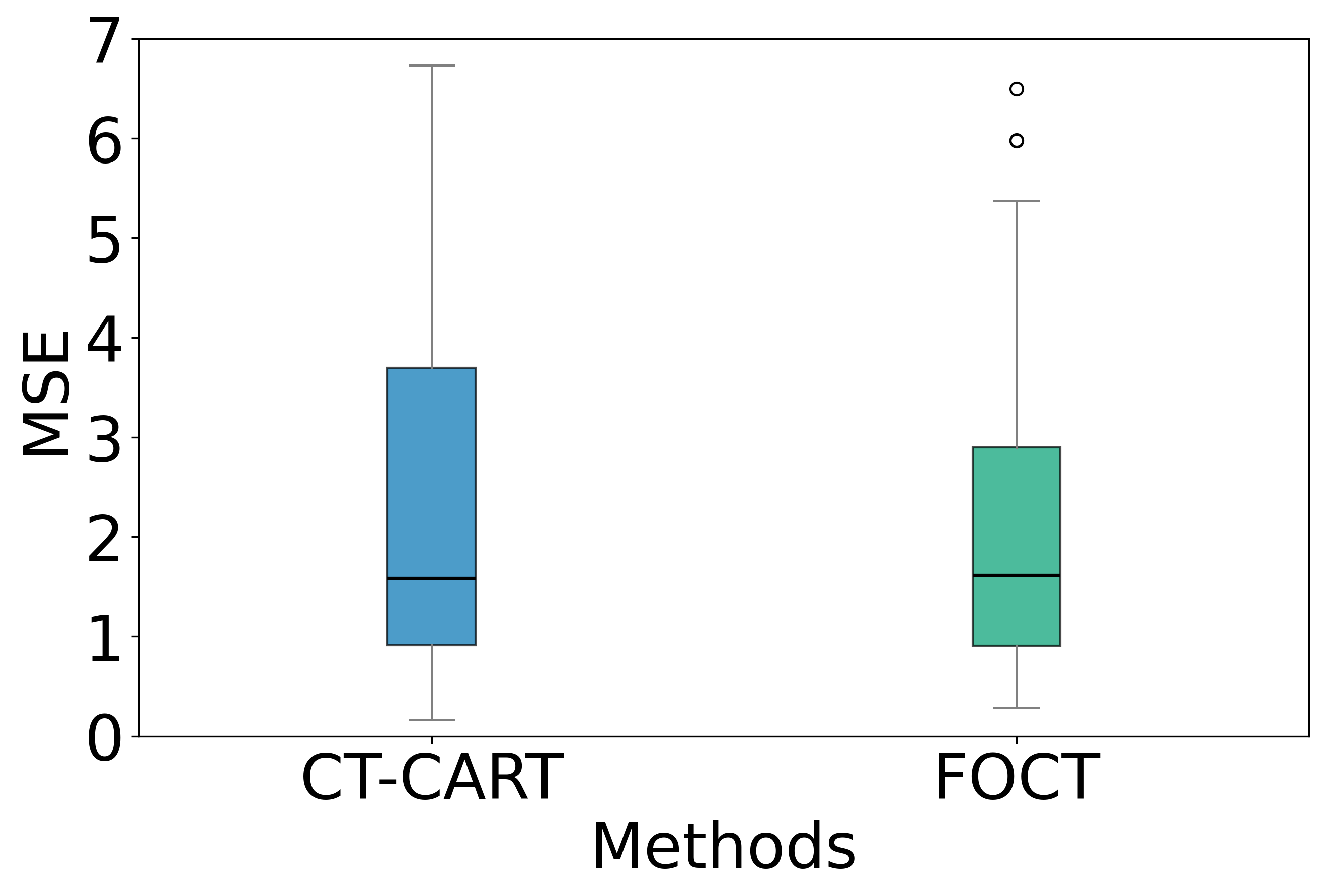}
        \caption{$n=100$, $p=0.3$, $\rho=0.5$ and $d=5$}
        \label{fig:saten100d5two}
    \end{subfigure}
    \hfill
    \begin{subfigure}{0.48\textwidth}       \includegraphics[width=\linewidth]{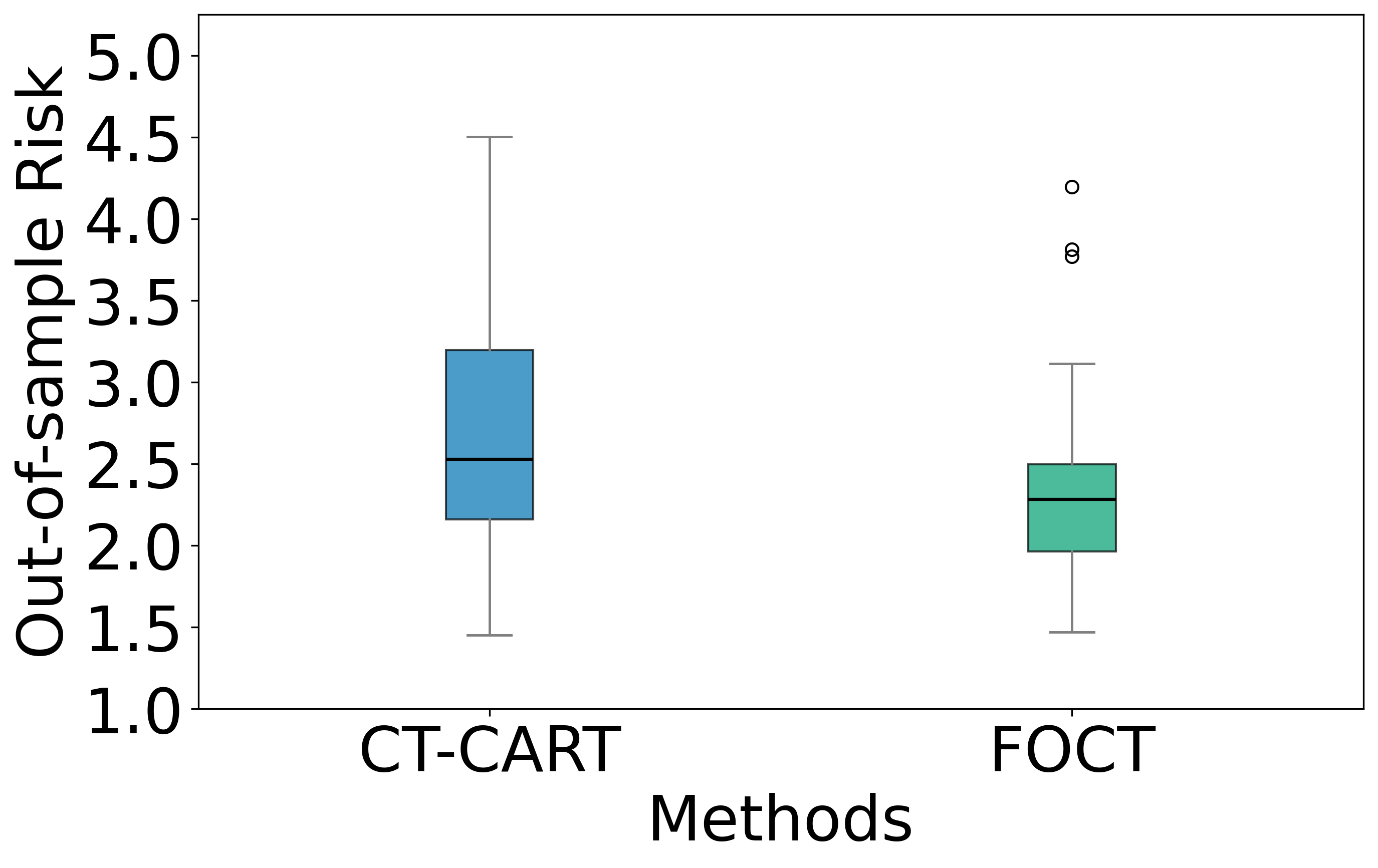}
        \caption{$n=100$, $p=0.3$, $\rho=0.5$ and $d=5$}
        \label{fig:riskn100d5two}
    \end{subfigure}
    \vspace{0.3in}
    \caption{Boxplot of mean square error (MSE) of SATE estimation and out-of-sample risk for CT-CART and FOCT with $n=100$, $p=0.3$, $\rho=0.5$ and $d=5$ across 50 Monte Carlo simulations.}
\end{figure}

We also explore a setting where we introduce additional irrelevant predictors to increase the complexity of subgroup identification and SATE estimation. Specifically, we add two independent, standard normal variables \(X_4\) and \(X_5\) into the leaf-level regression model, extending it to \(\,Y \sim [T,X_1,X_2,X_3,X_4,X_5,TX_1,TX_2,TX_3,TX_4,TX_5]\) while keeping the data-generating process unchanged. Including irrelevant covariates intensifies the overfitting problem, as demonstrated by results from 50 Monte Carlo simulations: CT-CART detects the correct tree structure only 9 times, OCT 6 times, and FOCT 22 times. A similar trend emerges for SATE estimation precision and out-of-sample risk, mirroring the findings in Section \ref{n=100,d=3} (Figure \ref{fig:saten100d5} and Figure \ref{fig:riskn100d5}). OCT, which relies purely on training data to find the optimal tree structure, suffers most from overfitting and thus exhibits poor out-of-sample performance. In contrast, FOCT’s fusion constraint mitigates overfitting, yielding superior performance on both SATE estimation and out-of-sample risk. Notably, FOCT also demonstrates a slight edge over CT-CART in these measures (Figure \ref{fig:saten100d5two} and Figure \ref{fig:riskn100d5two}).

\clearpage

\subsection{Balanced treatment and control groups with $p=0.5$ and smaller correlation between $X_2$ and $X_3$: $\rho=0.5$}

In this scenario, we set \(p=0.5\) in the equation \(T \sim \text{Bernoulli}(p)\) to generate balanced treatment and control groups. We also set the parameter \(\rho\) to 0.5, thereby reducing the correlation between \(X_2\) and \(X_3\). This setup mimics conditions with lower overfitting levels and decreased correlation between an actual split variable (\(X_2\)) and a spurious split variable (\(X_3\)). Consequently, we expect all three methods to exhibit improved performance. To compare their performances, we evaluate subgroup identification accuracy, SATE estimation precision, and out-of-sample risk across 50 Monte Carlo simulations. We tune the $\lambda$ in the grid $
\Lambda=\{1/10000*i|i=1,\ldots,20\} 
$

\begin{figure}[h]
    \centering
    \begin{subfigure}{0.46\textwidth}
      \includegraphics[width=\linewidth]{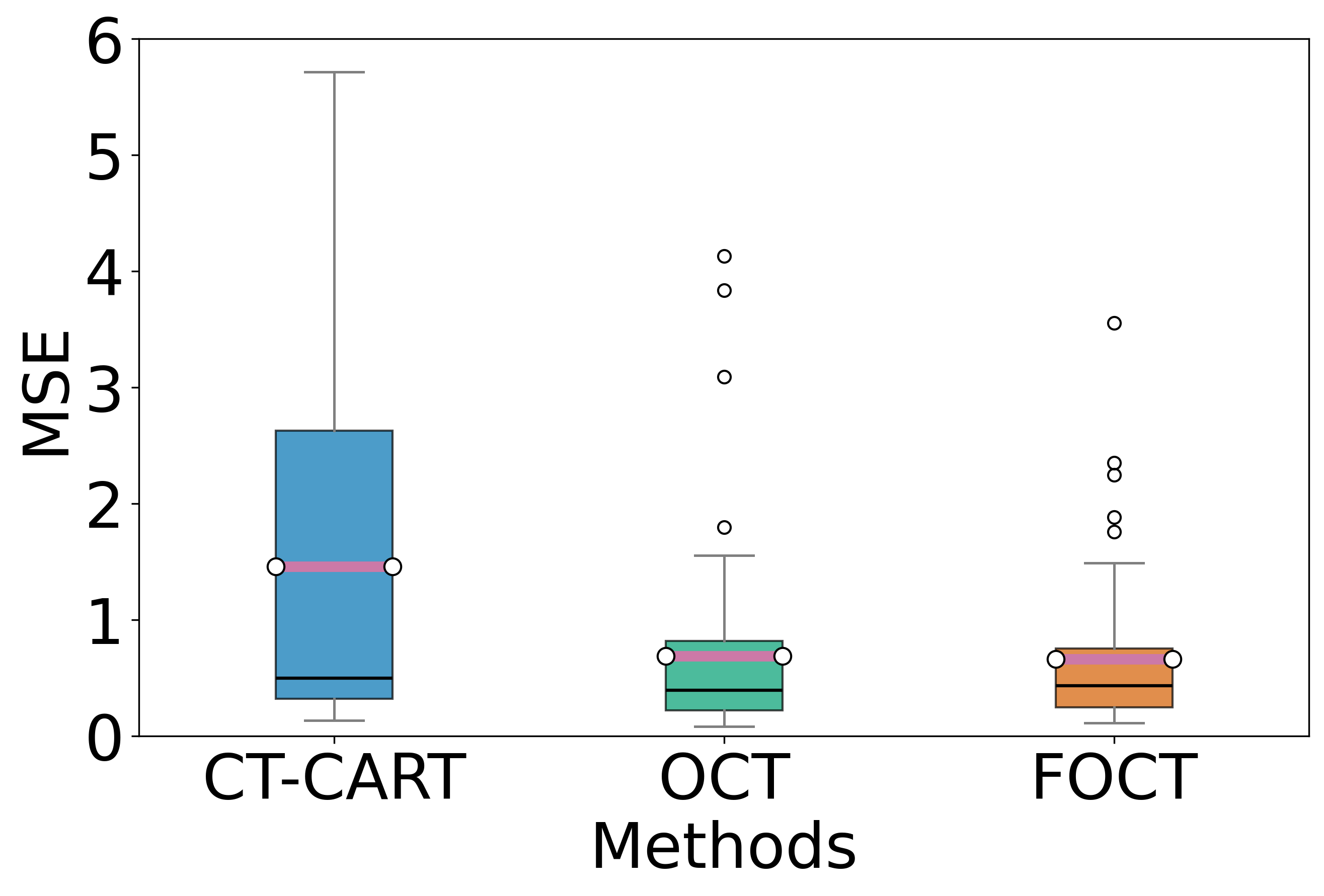}
        \caption{$n=200$, $\rho=0.5$, $p=0.5$ and $d=3$}
        \label{fig:satep0.5}
    \end{subfigure}
    \hfill
    \begin{subfigure}{0.48\textwidth}       \includegraphics[width=\linewidth]{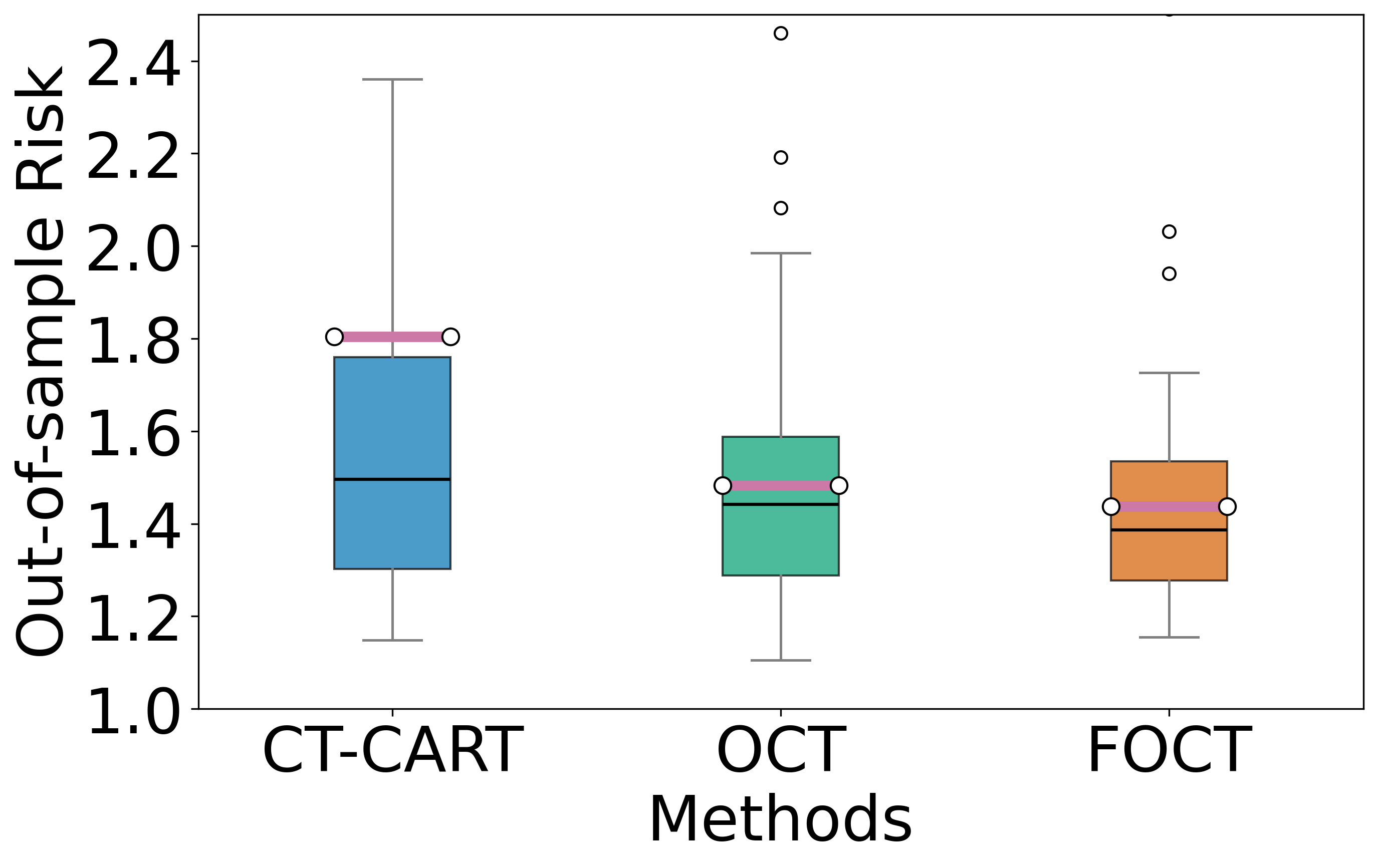}
        \caption{$n=200$, $\rho=0.5$, $p=0.5$ and $d=3$}
        \label{fig:riskp0.5}
    \end{subfigure}
    \vspace{0.3in}
    \caption{Boxplot of mean square error (MSE) of SATE estimation and out-of-sample risk for different methods with $\rho=0.5$, $p=0.5$, $n=200$, and $d=3$ across 50 Monte Carlo simulations. The purple line is the mean of MSE (or out-of-sample risk) across 50 Monte Carlo simulations for each method.}
\end{figure}

In this scenario, CT-CART identifies the correct tree structure 37 times, whereas OCT does so 46 times and FOCT 48 times. The superior performance of OCT and FOCT stems from their ability to yield optimal tree structures. Regarding SATE estimation precision and out-of-sample risk, OCT/FOCT again outperforms CT-CART, as reflected by smaller and more stable mean squared errors (MSEs) and out-of-sample risks (Figures \ref{fig:satep0.5} and \ref{fig:riskp0.5}). However, the performance gap between OCT and FOCT narrows in this setting, because the overfitting issue is less pronounced.

\subsection{Smaller correlation $\rho=0.6$ between $X_2$ and $X_3$}

We let $\rho=0.6$ and fix other parameters in the setup of the main paper. Then we compare the performances of different methods in terms of SATE estimation precision and out-of-sample risk. 

\begin{figure}[h]
    \centering
    \begin{subfigure}{0.48\textwidth}
      \includegraphics[width=\linewidth]{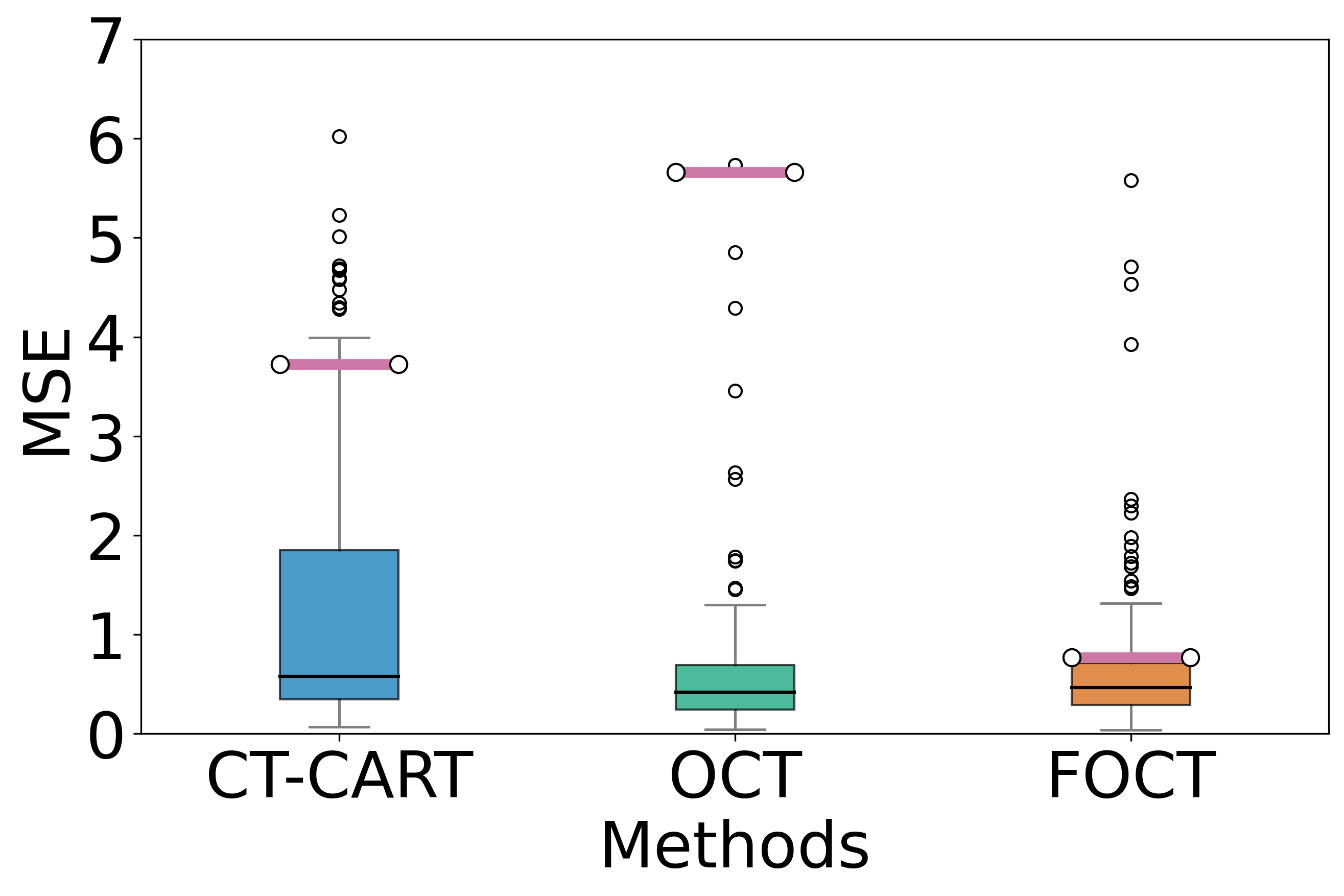}
        \caption{$n=200$, $\rho=0.6$, $p=0.3$ and $d=3$}
        \label{fig:sate0.6}
    \end{subfigure}
    \hfill
    \begin{subfigure}{0.48\textwidth}       \includegraphics[width=\linewidth]{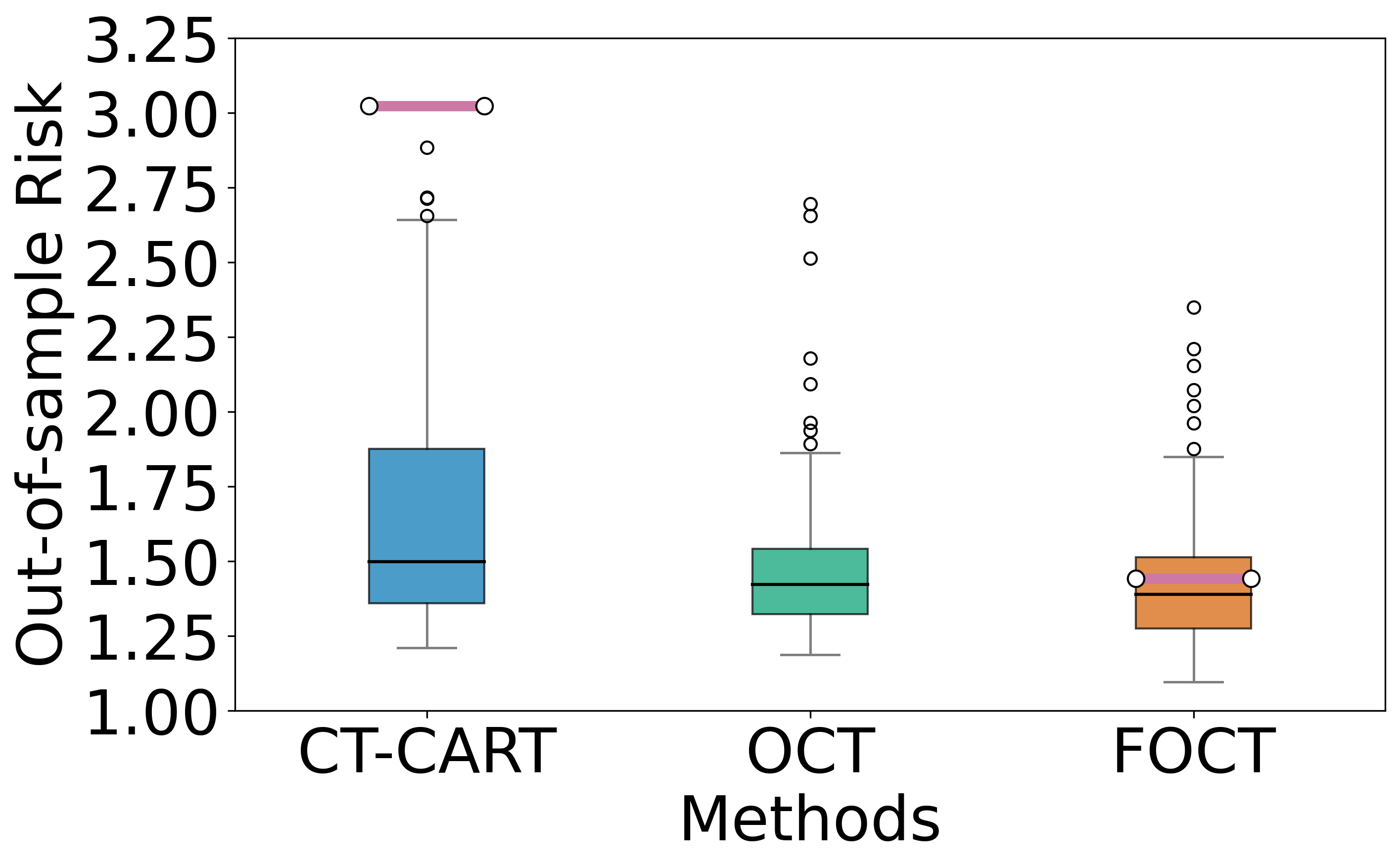}
        \caption{$n=200$, $\rho=0.6$, $p=0.3$ and $d=3$}
        \label{fig:risk0.6}
    \end{subfigure}
    \vspace{0.3in}
    \caption{Boxplot of mean square error (MSE) of SATE estimation and out-of-sample risk for different methods with $\rho=0.6$ across 100 Monte Carlo simulations. The purple line is the mean of MSE (or out-of-sample risk) across 100 Monte Carlo simulations for each method. The purple line for OCT in Figure \ref{fig:risk0.6} is above 3.25 and not shown.}
\end{figure}

With a smaller $\rho$, these three methods become more powerful in detecting the correct tree structures: FOCT (87 times),  OCT (82 times), and CT-CART (65 times). In terms of SATE estimation precision and out-of-sample risk, FOCT still performs better than the other two methods, with smaller and more stable MSE and out-of-sample risk, as shown in Figure \ref{fig:sate0.6} and Figure \ref{fig:risk0.6}.

\clearpage

\subsection{Discussion and comparison of additional baselines}

Within tree-based methods, Branch-and-Bound methods \cite{mazumder2022quant} and dynamic programming methods \cite{demirovic2022murtree} provide non-greedy optimal trees with piecewise constant leaves. However, these methods typically rely on top-down strategies that are incompatible with our global parameter fusion constraints — constraints that are essential in small-sample settings to reduce variance and avoid overfitting. As a result, they can not fully achieve our objectives. We hope to clarify that parameter fusion is not a minor modification in our method. It fundamentally alters the learned tree structure (subgroups)—a capability not present in existing tree-based methods, whether greedy or globally optimized. Failing to incorporate global parameter fusion is indeed a bottleneck of existing tree methods, as analyzed in \cite{tan2022cautionary}. Our simulations strongly support the benefit of fusion: Across all designs, FOCT consistently achieves higher subgroup recovery accuracy, demonstrating that fusion is essential for stable and reliable subgroup identification.

Beyond trees, clustering methods (e.g., 
k-means or Gaussian mixture models) are also used for subgroup discovery. However, the lack of interpretability makes them less suitable for our goals compared to tree-based methods: Traditional clustering methods group individuals based on some similarity metrics, but the resulting clusters are not defined by explicit, interpretable rules. This makes it difficult to translate the cluster assignments into actionable subgroup definitions for clinicians. With clustering methods, assigning new patients to a subgroup is not straightforward without retraining or embedding the new point. In contrast, tree-based methods provide clear, rule-based partitions of the covariate space (e.g., $Age > 70$ and $APOE4 = 1$), which can be directly used for clinical decision-making and scientific interpretation.

\subsubsection{Comparison with causal tree} Our goal is to identify interpretable subgroups through an explicit partition of the covariate space. A closely related method is the Causal Tree \citep{athey2016recursive}. 
To evaluate the performance of the causal tree \citep{athey2016recursive} in our context, we added a supplementary simulation under the scenario described in the main paper (with correlation $\rho=0.8$). We found that the causal tree correctly recovered the true subgroup structure (i.e., with split variable $X_1$ or $X_2$ at the root and the other at both child nodes) in only 3 out of 100 Monte Carlo replications. We believe this underperformance stems from causal trees’ reliance on sample splitting for honest estimation, which reduces the effective sample size available for tree construction and makes subgroup discovery even more challenging in small-sample settings. In contrast, our FOCT method leverages parameter fusion constraints to more accurately recover the underlying subgroup structure, as demonstrated in Section 5.4 of the main paper. This advantage arises from the ability of our MIO formulation to treat tree construction as a single global optimization problem, without the need for data splitting. Nevertheless, it would be interesting to further explore the connection between MIO-based optimization and sample-splitting-based causal trees in future work.

\subsubsection{Comparison with non-greedy optimal-tree methods} We also compared FOCT with an adapted version of the branch-and-bound (BnB) optimal-tree method with linear models in the leaves, as this is the only other optimal tree approach that directly accommodates both regression and continuous covariates. In our main setting with $n=200$, BnB recovered the true structure 63/100 times when $\rho=0.7$
 and 73/100 times when $\rho=0.8$
, whereas FOCT recovered it 67/100 and 81/100 times, respectively. BnB’s SATE MSE and out-of-sample risk closely resemble those of OCT and remain worse than FOCT. Under more challenging small-sample conditions (Appendix C.1.1), BnB recovered the true structure only 20/50 times, while FOCT recovered it 39/50 times. In this regime, BnB shows extremely poor SATE and risk performance, similar to OCT (Figure 5).

\subsubsection{Comparison with clustering methods} We added additional simulations on comparing k-means clustering (kmeans() function in R) and Gaussian mixture model clustering (mclust package in R) with the same linear regression model applied on the learned clusters. The data-generating process is the same as the main paper with $\rho=0.8$. We found that the overall out-of-sample risk of both methods is higher than the methods we compared in the paper (See Figure 1d) across the same 100 Monte Carlo simulations (1.57-3.21 for k-means and 1.82-8.68 for Gaussian mixture model clustering after removing 5 significant large outliers).

\newpage

\section{An additional assumption for the theoretical results of CART}\label{app:assumption}
 \begin{assumption}\label{assum4}
    Let $f^{*}(\cdot) \in \mathcal{F}^{h}$ and $K \geq h$. For any leaf node $\Tilde{l}_t$ of the depth $K-h$ tree $T_{K-h}$ such that $\widehat{\mathcal{R}}_{\Tilde{l}_t}\left(\hat{f}\left(T_{K-h}\right)\right)>\widehat{\mathcal{R}}_{\Tilde{l}_t}\left(f^{*}\right)$, we have
$$
\mathcal{I} \mathcal{G}_{h}(\Tilde{l}_t) \geq \frac{\left(\widehat{\mathcal{R}}_{\Tilde{l}_t}\left(\hat{f}\left(T_{K-h}\right)\right)-\widehat{\mathcal{R}}_{\Tilde{l}_t}\left(f^{*}\right)\right)^{2}}{V^{2}\left(f^{*}\right)},
$$
for some complexity constant $V\left(f^{*}\right)$ that depends only on $f^{*}(\cdot)$. Here
$$
\widehat{\mathcal{R}}_{\Tilde{l}_t}\left(\hat{f}\left(T_{K-h}\right)\right)=\frac{1}{|\Tilde{l}_t|}
\sum_{i \in \Tilde{l}_t} |Y_i-\hat{f}\left(T_{K-h}\right)(\boldsymbol{Z}_i)|^2 ,
$$
$$
\widehat{\mathcal{R}}_{\Tilde{l}_t}\left(f^*\right)=\frac{1}{|\Tilde{l}_t|}
\sum_{i \in \Tilde{l}_t} |Y_i-f^*(\boldsymbol{Z}_i)|^2,
$$
and $\mathcal{I} \mathcal{G}_{h}(\Tilde{l}_t)$ is the decrease in empirical risk from
greedily splitting $h$ times in the node $\Tilde{l}_t$.
\end{assumption}
\begin{remark}
    Assumption \ref{assum4} is a technical condition required for the theoretical analysis of the greedy tree algorithm. It requires that any $h$ sequential splits in the greedy tree must achieve a sufficient reduction in empirical risk. This assumption is introduced to address the technical challenges arising from the greedy structure of the tree algorithm. However, it is often difficult to verify because it depends on data-dependent quantities. Despite this stringent assumption, the greedy-based tree algorithm still exhibits a looser risk bound compared to the optimal tree.
\end{remark}

\section{Proof of the Theorectical Results}

In this section, we present the proofs of our theoretical findings in the main paper. We begin with introducing the notations and definitions that will be used throughout. Then we revisit the theorems in the main paper. Finally, we provide a detailed proof of each theorem.

\subsection{Notations and Definitions}

We first introduce several notations that will be used for the proof, which are cited and summarized from Chapter 9.4 and Chapter 13.1 of \cite{gyorfi2006distribution}.

We let $\Pi_{N}$ be the family of all achievable partitions $\mathcal{P}$ by growing a depth $K$ tree on $N$ points $\{(\boldsymbol{Z}_i,Y_i)\}_{i=1}^N$ and
$
M\left(\Pi_{N}\right):=\max \left\{ |\mathcal{P}|: \mathcal{P} \in \Pi_{N}\right\}
$
be the maximum number of terminal nodes among all partitions in $\Pi_{N}$. 
Given a set $\mathbf{Z}^{N}=\left\{\mathbf{Z}_{1}, \mathbf{Z}_{2}, \ldots, \mathbf{Z}_{N}\right\} \subset \mathbb{R}^{p}$, we define $\Delta\left(\mathbf{Z}^{N}, \Pi_{N}\right)$ to be the number of different partitions $\left\{\mathbf{Z}^{N} \cap A: A \in \mathcal{P}\right\}$, for $\mathcal{P} \in \Pi_{N}$. The partitioning number $\Delta_{N}\left(\Pi_{N}\right)$ is defined by
$$
\Delta_{N}\left(\Pi_{N}\right):=\max \left\{\Delta\left(\mathbf{Z}^{N}, \Pi_{N}\right): \mathbf{Z}_{1}, \mathbf{Z}_{2}, \ldots, \mathbf{Z}_{N} \in \mathbb{R}^{p}\right\}
$$
which is the maximum number of different partitions of any $N$ point set induced by elements of $\Pi_{N}$.
\begin{definition}[Shatter coefficient]
   Let $\mathcal{A}$ be a class of subsets of $\mathbb{R}^p$ and let $n \in \mathcal{N}$.

(a) For $z_{1}, \ldots, z_{n} \in \mathbb{R}^p$ define
$$
s\left(\mathcal{A},\left\{z_{1}, \ldots, z_{n}\right\}\right)=\left|\left\{A \cap\left\{z_{1}, \ldots, z_{n}\right\}: A \in \mathcal{A}\right\}\right|
$$
that is, $s\left(\mathcal{A},\left\{z_{1}, \ldots, z_{n}\right\}\right)$ is the number of different subsets of $\left\{z_{1}, \ldots, z_{n}\right\}$ of the form $A \cap\left\{z_{1}, \ldots, z_{n}\right\}, A \in \mathcal{A}$.

(b) Let $G$ be a subset of $\mathbb{R}^p$ of size $n$. One says that $\mathcal{A}$ shatters $G$ if $s(\mathcal{A}, G)=2^{n}$, i.e., if each subset of $G$ can be represented in the form $A \cap G$ for some $A \in \mathcal{A}$.

(c) The nth shatter coefficient of $\mathcal{A}$ is
$$
S(\mathcal{A}, n)=\max _{\left\{z_{1}, \ldots, z_{n}\right\} \subseteq \mathbb{R}^p} s\left(\mathcal{A},\left\{z_{1}, \ldots, z_{n}\right\}\right)
$$

That is, the shatter coefficient is the maximal number of different subsets of $n$ points that can be picked out by sets from $\mathcal{A}$. 
\end{definition}

\begin{definition}[Vapnik-Chervonenkis dimension]
Let $\mathcal{A}$ be a class of subsets of $\mathbb{R}^p$ with $\mathcal{A} \neq \emptyset$. The VC dimension (or Vapnik-Chervonenkis dimension) $V_{\mathcal{A}}$ of $\mathcal{A}$ is defined by
$$
V_{\mathcal{A}}=\sup \left\{n \in \mathcal{N}: S(\mathcal{A}, n)=2^{n}\right\}
$$
i.e., the $V C$ dimension $V_{\mathcal{A}}$ is the largest integer $n$ such that there exists a set of $n$ points in $\mathbb{R}^p$ which can be shattered by $\mathcal{A}$.    
\end{definition}

\begin{definition}[$\mathcal{F}^{+}$]
    We denote
$$
\mathcal{F}^{+}:=\left\{\left\{(z, t) \in \mathbb{R}^p \times \mathbb{R} ; t \leq f(z)\right\} ; f \in \mathcal{F}\right\}
$$
of all subgraphs of functions of $\mathcal{F}$.
\end{definition}

\subsection{Lemmas}
Before presenting the proof, we list some useful lemmas that will be used in the proof. The first four lemmas are cited from Theorem 9.4, Theorem 9.5, Theorem 11.4, and Lemma 13.1 of \cite{gyorfi2006distribution}. Lemma \ref{lem5} is cited from the proof of Theorem 4.3 in \cite{klusowski2024large}. 
\begin{lemma}\label{lem1}
 Let $\Pi$ be a family of partitions of $\mathbb{R}^p$ and let $\mathcal{F}$ be a class of function $f: \mathbb{R}^p \rightarrow \mathbb{R}$. Then one has for each $z_1,\ldots,z_n \in \mathbb{R}^p$ for each $\epsilon>0$, we have 
    \begin{align*}
        \mathcal{N}_1(\epsilon,\mathcal{F}\circ\Pi,z_1^n)\leq\Delta(z_1^n,\Pi)\left\{\sup_{x_1,\ldots,x_m\in z_1^n, m\leq n}\mathcal{N}_1(\epsilon,\mathcal{F}\circ\Pi,x_1^m)\right\}^{M(\Pi)}.
    \end{align*}
\end{lemma}
\vspace{0.2em}
\begin{lemma}\label{lem2}
Assume $|Y| \leq U_0$ a.s. and $U_0 \geq 1$. Let $\mathcal{F}$ be a set of functions $f: \mathbb{R}^p \rightarrow \mathbb{R}$ and let $|f(z)| \leq U_0, U_0 \geq 1$. Then, for each $n \geq 1$,
\begin{align*}
& \mathbf{P} \left\{ \exists f \in \mathcal{F} : \mathbb{E}|f(Z) - Y|^2 - \mathbb{E}|f^*(Z) - Y|^2 - \frac{1}{n} \sum_{i=1}^{n} \left( |f(Z_i) - Y_i|^2 - |f^*(Z_i) - Y_i|^2 \right) \right. \\
& \quad \left. \geq \epsilon \cdot \left( \alpha + \beta + \mathbb{E}|f(Z) - Y|^2 - \mathbb{E}|f^*(Z) - Y|^2 \right) \right\} \\&\leq 14 \sup_{Z_1^n} \mathcal{N}_1 \left( \frac{\beta \epsilon}{20 U_0}, \mathcal{F}, Z_1^n \right) \exp \left( -\frac{\epsilon^2 (1 - \epsilon) \alpha n}{214 (1 + \epsilon) U_0^4} \right),
\end{align*}
where $\alpha, \beta>0$ and $0<\epsilon \leq 1 / 2$.
\end{lemma}
\vspace{0.2em}

\begin{lemma}\label{lem3}
    Let $\mathcal{F}$ be an $r$-dimensional vector space of real functions on $\mathbb{R}^p$, and set
$$
\mathcal{A}=\{\{\boldsymbol{Z}: f(\boldsymbol{Z}) \geq 0\}: f \in \mathcal{F}\}
$$

Then
$$
V_{\mathcal{A}} \leq r.
$$
\end{lemma}
\vspace{0.2em}
\begin{lemma}\label{lem4}
Let $\mathcal{F}$ be a class of functions $f: \mathbb{R}^p \rightarrow[0, U_0]$ with $V_{\mathcal{F}+} \geq 2$, let $\nu$ be a probability measure on $\mathbb{R}^p$, and let $0<\epsilon<\frac{U_0}{4}$. Then
$$
\mathcal{N}\left(\epsilon, \mathcal{F},\|\cdot\|_{L_{1}(\nu)}\right)\leq \mathcal{M}\left(\epsilon, \mathcal{F},\|\cdot\|_{L_{1}(\nu)}\right) \leq 3\left(\frac{2 e U_0}{\epsilon} \log \frac{3 e U_0}{\epsilon}\right)^{V_{\mathcal{F}+}}.
$$
\end{lemma}
\vspace{0.2em}
\begin{lemma}\label{lem5}
\begin{equation*}
\Delta_{N}\left(\Pi_{N}\right) \leq((N-1) p)^{2^{K}-1} \leq(N p)^{2^{K}} 
\end{equation*}
for any depth $K$ tree partitions $\Pi_{N}$.
\end{lemma}

\subsection{Proof of Theorem \ref{thm:OT}}
To prove this theorem, we use a truncation technique and decompose \begin{align*}
\mathbb{E}_{\mathcal{D}_n}\left(\left\|\hat{f}\left(T_{K}^{\mathrm{OT}}\right)-f^{*}\right\|^{2}\right)= & \mathbb{E}_{\mathcal{D}_n}\left(\left\|\hat{f}\left(T_{K}^{\mathrm{OT}}\right)-f^{*}\right\|^{2} \cdot\mathbf{1}(E_U)\right)+\mathbb{E}_{\mathcal{D}_n}\left(\left\|\hat{f}\left(T_{K}^{\mathrm{OT}}\right)-f^{*}\right\|^{2} \cdot\mathbf{1}\left(E_U^{c}\right)\right). 
\end{align*}
where the event $E_U$ is defined as 
$$
E_U:=\left\{\left|Y_{i}\right| \leq U, i=1,2, \ldots, N\right\}.
$$
We first consider the first term $\mathbb{E}_{\mathcal{D}_n}\left(\left\|\hat{f}\left(T_{K}^{\mathrm{OT}}\right)-f^{*}\right\|^{2} \cdot\mathbf{1}(E_U)\right)$.

To bound this term, we can rewrite
$\left\|\hat{f}\left(T_{K}^{\text {OT }}\right)-f^{*}\right\|^{2}=E_{1}+E_{2}$, where

\begin{equation*}
E_{1}:=\left\|\hat{f}\left(T_{K}^{\mathrm{OT}}\right)-f^{*}\right\|^{2}-\frac{2}{|\mathcal{D}_n|}\left(
\sum_{i \in \mathcal{D}_n} |Y_i-\hat{f}\left(T_{K}^{\mathrm{OT}}\right)(\boldsymbol{Z}_i)|^2 -\sum_{i \in \mathcal{D}_n}|Y_i-f^*(\boldsymbol{Z}_i)|^2\right)-\alpha-\beta 
\end{equation*}

and $$
E_{2}:=\frac{2}{|\mathcal{D}_n|}\left(
\sum_{i \in \mathcal{D}_n} |Y_i-\hat{f}\left(T_{K}^{\mathrm{OT}}\right)(\boldsymbol{Z}_i)|^2 -\sum_{i \in \mathcal{D}_n}|Y_i-f^*(\boldsymbol{Z}_i)|^2\right)+\alpha+\beta
$$
Here $\alpha$ and $\beta$ are positive constants that we specify later. 

Since $T_{K}^{\mathrm{OT}}$ is an optimal tree and $f^* \in \mathcal{F}_h$ with the form 
$$
f^*(\boldsymbol{Z})=\sum_{j=1}^{M(T_K^*)}\mathrm{1}_{\{\boldsymbol{Z}\in \mathcal{A}_j^*\}}\boldsymbol{\gamma}^{*T}_j \boldsymbol{Z}.
 $$
we have
\begin{align*}
   \frac{2}{|\mathcal{D}_n|}\left(
\sum_{i \in \mathcal{D}_n} |Y_i-\hat{f}\left(T_{K}^{\mathrm{OT}}\right)(\boldsymbol{Z}_i)|^2 -\sum_{i \in \mathcal{D}_n}|Y_i-f^*(\boldsymbol{Z}_i)|^2\right)\leq 0 \quad \text{   and } \quad
E_{2} \leq \alpha+\beta.
\end{align*}
Thus
\begin{align*}
\mathbb{E}_{\mathcal{D}_n}\left(E_{2}\cdot\mathbf{1}(E_U)\right) \leq \mathbb{E}_{\mathcal{D}_n}\left(E_{2}\right) \leq \alpha+\beta .
\end{align*}
Now it remains to bound $E_{1}$. We let $\mathcal{F}_{N}$ denote the collection of all piece-wise linear functions (bounded by $U$) on partitions $\mathcal{P} \in \Pi_{N}$.

Now, by using Lemma \ref{lem2} and choosing $\epsilon=1 / 2$, we have
\begin{align*}
&\mathbb{P}_{\mathcal{D}_n}\left(E_{1} \geq 0\right) \\& \leq \mathbb{P}_{\mathcal{D}_n}\left(\exists f(\cdot) \in \mathcal{F}_{N}:\left\|f-f^{*}\right\|^{2} \geq \frac{2}{|\mathcal{D}_n|}\left(
\sum_{i \in \mathcal{D}_n} |Y_i-\hat{f}\left(T_{K}^{\mathrm{OT}}\right)(\boldsymbol{Z}_i)|^2 -\sum_{i \in \mathcal{D}_n}|Y_i-f^*(\boldsymbol{Z}_i)|^2\right)+\alpha+\beta\right) \\
& \leq 14 \sup _{\mathbf{Z}^{N}} \mathcal{N}\left(\frac{\beta}{40 U}, \mathcal{F}_{N}, L_{1}\left(\mathbb{P}_{\mathbf{Z}^{N}}\right)\right) \exp \left(-\frac{\alpha N}{2568 U^{4}}\right),
\end{align*}
where $\mathbf{Z}^{N}=\left\{\mathbf{Z}_{1}, \mathbf{Z}_{2}, \ldots, \mathbf{Z}_{N}\right\} \subset \mathbb{R}^{p}$ and $\mathcal{N}\left(r, \mathcal{F}_{N}, L_{1}\left(\mathbb{P}_{\mathbf{Z}^{N}}\right)\right)$ is the covering number for $\mathcal{F}_{N}$ by balls of radius $r>0$ in $L_{1}\left(\mathbb{P}_{\mathbf{Z}^{N}}\right)$ with respect to the empirical discrete measure $\mathbb{P}_{\mathbf{Z}^{N}}$ on $\mathbf{Z}^{N}$.

By Lemma \ref{lem3}, we have the following result:
If $\mathcal{F}$ is a class of linear functions on $\mathbb{R}^p$, we have 
$$
V_{\mathcal{F}^+}\leq p+1.
$$
We note that the collection of all piece-wise linear functions $\mathcal{F}_N$ can be written as $\mathcal{F}\circ\Pi_N$ as in Lemma \ref{lem1}. Thus by Lemma \ref{lem1}, we can show 
    \begin{align*}
\mathcal{N}\left(\frac{\beta}{40 U}, \mathcal{F}_N, L_{1}\left(\mathbb{P}_{\mathbf{Z}^{N}}\right)\right)\leq\Delta_{N}\left(\Pi_{N}\right)\left\{\sup_{\boldsymbol{X}_1,\ldots,\boldsymbol{X}_m\in \boldsymbol{Z}_1^n, m\leq n}\mathcal{N}(\frac{\beta}{40 U},\mathcal{F},L_{1}\left(\mathbb{P}_{\mathbf{X}_{1:m}}\right))\right\}^{M(\Pi)}
    \end{align*}

We now can use Lemma \ref{lem4} to bound the following empirical covering number by
\begin{align*}
\sup_{\boldsymbol{X}_1,\ldots,\boldsymbol{X}_m\in \boldsymbol{Z}_1^n, m\leq n}\mathcal{N}(\frac{\beta}{40 U},\mathcal{F},L_{1}\left(\mathbb{P}_{\mathbf{X}_{1:m}})\right)&\leq3\left(\frac{160 e U^2}{\beta} \log \frac{240 e U^2}{\beta}\right)^{V_{\mathcal{F}+}} \\&\leq 3\left( \frac{ 240 e U^{2}}{\beta}\right)^{2V_{\mathcal{F}+}}\\&\leq 3\left( \frac{ 240 e U^{2}}{\beta}\right)^{2(p+1)}.
\end{align*}
Then we can get
\begin{equation*}
\mathcal{N}\left(\frac{\beta}{40 U}, \mathcal{F}_{N}, L_{1}\left(\mathbb{P}_{\mathbf{Z}^{N}}\right)\right) \leq \Delta_{N}\left(\Pi_{N}\right)\left(3\left( \frac{ 240 e U^{2}}{\beta}\right)^{p+1}\right)^{2 M\left(\Pi_{N}\right)} \leq \Delta_{N}\left(\Pi_{N}\right)\left(\frac{C_0^{p+1} U^{2p+2}}{\beta^{p+1}}\right)^{2^{K+1}}.
\end{equation*}
Here $C_0$ is a large enough constant.

By Lemma \ref{lem5} we have
\begin{equation*}
\Delta_{N}\left(\Pi_{N}\right) \leq(N p)^{2^{K}}.
\end{equation*}

We therefore can bound the covering number by
\begin{equation*}
\mathcal{N}\left(\frac{\beta}{40 U}, \mathcal{F}_{N}, L_{1}\left(\mathbb{P}_{\mathbf{Z}^{N}}\right)\right) \leq(N p)^{2^{K}}\left(\frac{C_0^{p+1} U^{2p+2}}{\beta^{p+1}}\right)^{2^{K+1}}.
\end{equation*}

Since $\hat{f}\left(T_{K}^{\text {OT }}\right)(\cdot) \in$ $\mathcal{F}_{N}$, we have
$$
\mathbb{P}_{\mathcal{D}_n}\left(E_{1} \geq 0|E_U\right) \leq 14(N p)^{2^{K}}\left(\frac{C_0^{p+1} U^{2p+2}}{\beta^{p+1}}\right)^{2^{K+1}} \exp \left(-\frac{\alpha N}{2568 U^{4}}\right).
$$
If we choose 
$$\alpha = C_1' \cdot \frac{U^{4} 2^{K} \log (N^{p+1} p)}{N} \text{ and }
 \beta =C_2' \cdot \frac{U^{2}}{N}$$
with positive constants $C_1'$ and $C_2'$, then we have
\begin{align*}
    &\quad14(N p)^{2^{K}}\left(\frac{C_0^{p+1} U^{2p+2}}{\beta^{p+1}}\right)^{2^{K+1}} \exp \left(-\frac{\alpha N}{2568 U^{4}}\right)\\&\leq 14(N p)^{2^{K}}\left(\frac{C_0^{p+1} N^{p+1} U^{2p+2}}{C_2'^{p+1}\cdot U^{2p+2}}\right)^{2^{K+1}} \exp \left(-\frac{C_1'\cdot U^{4} 2^{K} \log (N^{p+1} p) }{2568 U^{4}}\right)\\&\leq 14(N p)^{2^{K}}\left(\frac{C_0^{p+1} N^{p+1}}{C_2'^{p+1}} \right)^{2^{K+1}} \exp \left(-\frac{ C_1'\cdot 2^{K} \log (N^{p+1} p) }{2568 }\right)\\& \leq 14N^{2^{K}} N ^{(p+1)\cdot 2^{K+1}}N^{-(p+1)\frac{C_1'\cdot2^{K}}{2568}}p^{2^{K}}\left(\frac{C_0^{p+1}}{C_2'^{p+1}} \right)^{2^{K+1}} p^{-\frac{C_1'\cdot2^{K}}{2568}}\\& =14N^{(2p+3)\cdot 2^{K}} N^{-(p+1)\frac{C_1'\cdot2^{K}}{2568}}p^{2^{K}}\left(\frac{C_0^{p+1}}{C_2'^{p+1}} \right)^{2^{K+1}} p^{-\frac{C_1'\cdot2^{K}}{2568}}.
\end{align*}
If we further let $C_1'$ be large enough so that $C_1'/2568\geq 3$ and $C_2'\geq C_0$, then we are able to show 
\begin{align*}
  14N^{(2p+3)\cdot 2^{K}} N^{-(p+1)\frac{C_1'\cdot2^{K}}{2568}}p^{2^{K}}\left(\frac{C_0^{p+1}}{C_2'^{p+1}} \right)^{2^{K+1}} p^{-\frac{C_1'\cdot2^{K}}{2568}}\leq  C' N^{-2^{K}}\leq C'/N.
\end{align*}
for some universal constant $C^{\prime}$. Therefore, $\mathbb{P}_{\mathcal{D}_n}\left(E_{1} \geq 0|E_U\right) \leq C^{\prime} / N$. 

Furthermore, we have $E_{1} \leq\left\|\hat{f}\left(T_{K}^{\text {OT}}\right)-f^{*}\right\|^{2}+2\left\|Y-f^{*}\right\|_{\mathcal{D}_n}^{2}$. Condition on the event $E_U$, 
$$
E_{1} \leq\left\|\hat{f}\left(T_{K}^{\text {OT}}\right)-f^{*}\right\|^{2}+2\left\|Y-f^{*}\right\|_{\mathcal{D}_n}^{2}\leq 12 U^{2}.
$$
Thus,
$$\mathbb{E}_{\mathcal{D}_n}\left(E_{1}\cdot\mathbf{1}(E_U)\right) \leq 12 U^{2} \mathbb{P}_{\mathcal{D}_n}\left(E_{1} \geq 0|E_U\right) \leq 12 C^{\prime} U^{2} / N.$$

With the choices of $\alpha$ and $\beta$, we then have
\begin{equation}\label{eq1}
 \begin{aligned}
\mathbb{E}_{\mathcal{D}_n}\left(\left\|\hat{f}\left(T_{K}^{\mathrm{OT}}\right)-f^{*}\right\|^{2} \cdot\mathbf{1}(E_U)\right)& =\mathbb{E}_{\mathcal{D}_n}\left(E_{1}\cdot\mathbf{1}(E_U)\right)+\mathbb{E}_{\mathcal{D}_n}\left(E_{2}\cdot\mathbf{1}(E_U)\right) \\&\leq C^{\prime \prime}\left(\frac{U^{4} 2^{K} \log (N^{p+1} p)}{N}+\frac{U^{2}}{N}\right)
\end{aligned}   
\end{equation}
where $C^{\prime \prime}$ is a positive universal constant.

Next, we consider the second term $\mathbb{E}_{\mathcal{D}_n}\left(\left\|\hat{f}\left(T_{K}^{\mathrm{OT}}\right)-f^{*}\right\|^{2} \cdot\mathbf{1}\left(E_U^{c}\right)\right)$. If we choose $U$ such that $U \geq\left\|f^{*}\right\|_{\infty}$, we will have
\begin{equation*}
\mathbb{P}_{\mathcal{D}_n}\left(E_U^{c}\right)=\mathbb{P}(\cup_{i=1}^N|Y_i|>U)\leq N \mathbb{P}(|y_1|>U) \leq N \mathbb{P}\left(|\varepsilon|>U-\left\|f^{*}\right\|_{\infty}\right) \leq 2 N \exp \left(-\frac{\left(U-\left\|f^{*}\right\|_{\infty}\right)^{2}}{2 \sigma^{2}}\right).
\end{equation*}
Since we have
\begin{align*}
\left\|\hat{f}\left(T_{K}^{\mathrm{OT}}\right)\right\|_\infty= \sup_{\boldsymbol{Z} \in \boldsymbol{\mathcal{Z}}} |\sum_{\Tilde{l}_t \in \mathcal{T}_L(T_{K}^{\mathrm{OT}})}\mathrm{1}_{(\boldsymbol{Z}\in \Tilde{l}_t)}(\hat{\boldsymbol{\gamma}}_{\Tilde{l}_t})^T \boldsymbol{Z}|\leq \sup_{\Tilde{l}_t \in \mathcal{T}_L(T_{K}^{\mathrm{OT}})}\|\hat{\boldsymbol{\gamma}}_{\Tilde{l}_t}\|_1\leq B,
\end{align*}
and
\begin{align*}
\left\|f^*\right\|_\infty= \sup_{\boldsymbol{Z} \in \boldsymbol{\mathcal{Z}}} |\sum_{\Tilde{l}_t \in \mathcal{T}_L(T_K^*)}\mathrm{1}_{\{\boldsymbol{Z}\in \Tilde{l}_t\}}\boldsymbol{\gamma}_{\Tilde{l}_t}^{*T} \boldsymbol{Z}\|\leq \sup_{\Tilde{l}_t \in \mathcal{T}_L(T_{K}^{*})}\|\boldsymbol{\gamma}^*_{\Tilde{l}_t}\|_1\leq B,
\end{align*}
We now have
\begin{align*}
\mathbb{E}_{\mathcal{D}_n}\left(\left\|\hat{f}\left(T_{K}^{\mathrm{OT}}\right)-f^{*}\right\|^{2} \mathbf{1}\left(E_U^{c}\right)\right) & \leq 4 B^2\mathbb{P}_{\mathcal{D}_n}\left(E_U^{c}\right)\leq 8 B^2 N \exp \left(-\frac{\left(U-\left\|f^{*}\right\|_{\infty}\right)^{2}}{2 \sigma^{2}}\right).
\end{align*}
Choosing $U=B+\sqrt{2} \sigma \sqrt{2 \log (N)}$ yields 
$$
8 B^2 N \exp \left(-\frac{\left(U-\left\|f^{*}\right\|_{\infty}\right)^{2}}{2 \sigma^{2}}\right) \leq 
8 B^2 N \exp \left(-\frac{\left(U-B\right)^{2}}{2 \sigma^{2}}\right) = 8 B^2 N \exp \left(- 2 \log(N)\right)=\frac{8B^2}{N}.
$$
and therefore we have
$$\mathbb{E}_{\mathcal{D}_n}\left(\left\|\hat{f}\left(T_{K}^{\text {OT}}\right)-f^{*}\right\|^{2} \mathbf{1}\left(E_U^{c}\right)\right) \leq \frac{8B^2}{N}.$$ 
With this choice of $U$,  we can combine the bound in Equation \eqref{eq1} and obtain
\begin{align*}
\mathbb{E}_{\mathcal{D}_n}\left(\left\|\hat{f}\left(T_{K}^{\mathrm{OT}}\right)-f^{*}\right\|^{2}\right) &\leq C^{\prime \prime}\left(\frac{U^{4} 2^{K} \log (N^{p+1} p)}{N}+\frac{U^{2}}{N}\right)+\frac{8B^2}{N} \\&\leq C_{1} \frac{2^{K} \log^2(N) \big((p+1)\cdot\log (N)+\log(p)\big)}{N}.
\end{align*}
for some constant $C_{1}>0$ that depends only on $B$ and $\sigma^{2}$.

\subsection{Proof of Theorem \ref{thm:CART}}
The proof of Theorem \ref{thm:CART} is similar to the proof of Theorem \ref{thm:OT}. The main difference is the bound on $E_2$.  
$$
E_{2}:=\frac{2}{|\mathcal{D}_n|}\left(
\sum_{i \in \mathcal{D}_n} |Y_i-\hat{f}\left(T_{K}^{\mathrm{CART}}\right)(\boldsymbol{Z}_i)|^2 -\sum_{i \in \mathcal{D}_n}|Y_i-f^*(\boldsymbol{Z}_i)|^2\right)+\alpha+\beta
$$
with $T_{K}^{\mathrm{OT}}$ replaced by $T_{K}^{\mathrm{CART}}$.
We know that $T_{K}^{\mathrm{CART}}$ is a sub-optimal tree, so 
\begin{align*}
   \frac{2}{|\mathcal{D}_n|}\left(
\sum_{i \in \mathcal{D}_n} |Y_i-\hat{f}\left(T_{K}^{\mathrm{CART}}\right)(\boldsymbol{Z}_i)|^2 -\sum_{i \in \mathcal{D}_n}|Y_i-f^*(\boldsymbol{Z}_i)|^2\right)\leq 0 
\end{align*}
no longer holds. To bound this term, we need to introduce an additional assumption, which is Assumption \ref{assum4}.

With Assumption \ref{assum4}, following Lemma D.1 in \cite{klusowski2024large} we have  
$$\frac{1}{|\mathcal{D}_n|}\left(
\sum_{i \in \mathcal{D}_n} |Y_i-\hat{f}\left(T_{K}^{\mathrm{CART}}\right)(\boldsymbol{Z}_i)|^2 -\sum_{i \in \mathcal{D}_n}|Y_i-f^*(\boldsymbol{Z}_i)|^2\right)\leq \frac{V(f^*)h}{K+2h+1}.$$

Thus 
\begin{align*}
\mathbb{E}_{\mathcal{D}_n}\left(E_{2}\cdot\mathbf{1}(E_U)\right) \leq \mathbb{E}_{\mathcal{D}_n}\left(E_{2}\right) \leq \frac{2V(f^*)h}{K+2h+1}+\alpha+\beta .
\end{align*}

By using similar proof in Theorem \ref{thm:OT} with $T_{K}^{\mathrm{OT}}$ replaced by $T_{K}^{\mathrm{CART}}$, we have 

\begin{align*}
\mathbb{E}_{\mathcal{D}_n}\left(\left\|\hat{f}\left(T_{K}^{\mathrm{CART}}\right)-f^{*}\right\|^{2}\right) &\leq \frac{2V(f^*)h}{K+2h+1}+C^{\prime \prime}\left(\frac{U^{4} 2^{K} \log (N^{p+1} p)}{N}+\frac{U^{2}}{N}\right)+\frac{8B^2}{N} \\&\leq C_{1} \frac{2^{K} \log^2(N) \big((p+1)\cdot\log (N)+\log(p)\big)}{N}+\frac{2V(f^*)h}{K+2h+1}.
\end{align*}

Let $C_0=2V(f^*)$ then we finish the proof of Theorem \ref{thm:CART}.

\section{An Example that Wrong Split Produces Non-Vanishing Approximation Error.}\label{example}
Consider a one-depth tree, let the true split be at $c^* = 1/2$, and suppose the estimator splits at a fixed
$c \neq 1/2$.  
The true regression function is
\[
f^*(x)
=
\begin{cases}
a_1 x, & 0 \le x < 1/2,\\[4pt]
a_2 x, & 1/2 \le x \le 1,
\end{cases}
\qquad a_1 \neq a_2.
\]
The estimator uses the incorrect partition
\[
[0,c)
\quad\text{and}\quad
[c,1],
\qquad c \neq \tfrac12,
\]
and fits on each leaf the linear predictors
\[
g_1(x) = b_1 x, \qquad x \in [0,c),
\qquad\qquad
g_2(x) = b_2 x, \qquad x \in [c,1].
\]

Assume $c < 1/2$ (the case $c > 1/2$ is symmetric). The true regression function is
\[
f^*(x)
=
\begin{cases}
a_1 x, & c \le x < 1/2,\\[4pt]
a_2 x, & 1/2 \le x \le 1.
\end{cases}
\]

The best possible linear predictor on leaf 2 minimizes
\[
R_{\mathrm{leaf\,2}}
=
\min_{b_2}
\left\{
\int_c^{1/2} (b_2 - a_1)^2 x^2\,dx
+
\int_{1/2}^1 (b_2 - a_2)^2 x^2\,dx
\right\}.
\]
Define
\[
I_1(c)
:= 
\int_c^{1/2} x^2\,dx
=
\frac{1}{24} - \frac{c^3}{3},
\qquad
I_2
:=
\int_{1/2}^1 x^2\,dx
=
\frac{7}{24}.
\]
Then
\[
R_{\mathrm{leaf\,2}}
=
\min_{b_2}
\bigl[
I_1(c)(b_2 - a_1)^2 + I_2(b_2 - a_2)^2
\bigr].
\]
And it takes the minimum when
\[
b_2=b_2^*
=
\frac{I_1(c)a_1 + I_2 a_2}{\,I_1(c) + I_2\,}.
\]
And
\[
R_{\mathrm{leaf,2}}(b_2^*)
=
\frac{I_1(c) I_2}{I_1(c)+I_2}\,(a_1 - a_2)^2.
\]
If
\(
\bigl|c - \tfrac12\bigr| = \varepsilon>0
\)
for some fixed $\varepsilon>0$, then
\[
I_1(c)=\delta(\varepsilon) \;>\; 0,
\]
and
\[
R_{\mathrm{leaf,2}}(b_2^*)
\;\ge\;
\frac{\delta(\varepsilon)\,(7/24)}{\delta(\varepsilon)+7/24}\,(a_1 - a_2)^2
=
c_0(\varepsilon)\,(a_1 - a_2)^2
\;>\; 0,
\]
So for any fixed misaligned split $c \neq c^*$,
\[
\inf_{\hat f \text{ using split } c}
E\bigl[\|\hat f - f^*\|_2^2\bigr]
\;\ge\;
c_0(\varepsilon)\,(a_1 - a_2)^2
\;>\;
0.
\]
This lower bound does not depend on the sample size $n$. Hence an estimator achieving a minimax rate of order $O(n^{-1})$ must satisfy 
$c \to c^*$; any persistent mismatch in the split induces a non-vanishing
approximation error and prevents the estimator from attaining this rate.
\clearpage

\section{Mixed integer optimization framework}\label{app:mio}
MIO refers to the optimization problem involving both integer
variables (which can take on only whole number values) and continuous variables (which can take on any real
number values). The feasibility of this MIO-based optimal partition searching approach is bolstered by significant
algorithmic advancements in integer optimization and hardware enhancements over the past 25 years. The general mixed integer optimization problem takes the form 
\begin{align*}
    &\quad \min_{\boldsymbol{\alpha} \in \mathbb{R}^m} \boldsymbol{\alpha}^T \boldsymbol{Q}\boldsymbol{\alpha} + \boldsymbol{\alpha}^T \boldsymbol{u}
    \\& \text{s.t. } \boldsymbol{A}\boldsymbol{\alpha} \leq \boldsymbol{v},
    \\& \quad\ \   \alpha_j \in \{0, 1\},\quad j \in \mathcal{I},
    \\& \quad\ \  \alpha_j \geq 0, \quad\quad\quad j \notin \mathcal{I}.
\end{align*}
where $\boldsymbol{u} \in \mathbb{R}^m, \boldsymbol{A} \in \mathbb{R}^{h\times m}, \boldsymbol{v} \in \mathbb{R}^h$, and $\boldsymbol{Q} \in \mathbb{R}^{m\times m}$ is positive semi-definite. $\boldsymbol{\alpha}$ is the mixture of discrete and continuous components and $\mathcal{I}$ is used to identify the binary components of $\boldsymbol{\alpha}$. Any optimization problem that can be expressed in this form qualifies as a mixed integer optimization problem and can be solved using popular MIO solvers such as Gurobi \citep{gurobi}. 

\section{Detailed notations in Algorithm \ref{algo:mio}} \label{sec:notation}

\begin{itemize}
    \item We use $[n]$ to represent the integer from $1$ to $n$. 
    \item To mathematically represent the tree structure, we divide its nodes into two classes: \(\mathcal{T}_B\), the set of branch nodes where splits occur, and \(\mathcal{T}_L\), the collection of leaf nodes, each corresponding to a final identified subgroup. 
    \item  We denote the parent node of branch node $t'$ as $\texttt{p}(t')$ and the collection of its ancestors in the left (right) branch as $\mathcal{L}(t')$ ($\mathcal{R}(t')$). 
    \item  To allow for partially grown trees (where the number of leaf nodes is less than \(2^D\) when the tree depth is \(D\)), we introduce binary variables \(\{l_{t'}\}_{t' \in \mathcal{T}_L}\), with \(l_{t'}=0\) indicating that leaf node \(t'\) is empty.
    \item  We also introduce a binary vector \(\boldsymbol{a}_{t'}\in \{0,1\}^d\) and a variable \(b_{t'}\) to model the split of branch nodes \(t' \in \mathcal{T}_B\). Here \(\boldsymbol{a}_{t'}\) determine the split variable in $t' \in \mathcal{T}_B$ and $b_{t'}$ is the cutoff of this split variable. For example, \(a_{t'j}=1\) indicates that in branch node \(t'\), the sample is divided into two partitions: \(\{(\boldsymbol{X}_i, T_i, Y_i) \mid \boldsymbol{X}_i  \in t', X_{ij}<b_{t'}\}\) and  \(\{(\boldsymbol{X}_i, T_i, Y_i) \mid \boldsymbol{X}_i  \in t', X_{ij}\geq b_{t'}\}\).
    \item  To accommodate branch nodes with no split, we use binary variables \(d_{t'}\) with \(d_{t'}=0\) indicating no split occurs in branch node \(t' \in \mathcal{T}_B\). 
    \item To allow for the parameter fusion, we further introduce binary variables\(\{r_{j t_1t_2}\}_{j=1}^{2d+2}\) to indicate whether we fuse two parameters on the same covariates in different subpopulations, with value $r_{j t_1 t_2}=1$ indicating that $\gamma_{t_1j}=\gamma_{t_2j}$ and value $r_{j t_1 t_2}=0$ indicating no fusion constraints between $\gamma_{t_1j}$ and $\gamma_{t_2j}$. 
\end{itemize}

\clearpage

\section{Further details on Section \ref{sec:method}}\label{app:detail on tree}

\subsection{Regression trees and hierarchical partitions}
The tree is in one-to-one correspondence with a hierarchy partition \citep{chatterjee2021adaptive}.  A hierarchical partition is formed by recursively applying a series of hierarchical splits to the sample space \(\mathcal{A} = \Pi_{j=1}^{d} [a_j, b_j]\), where each split involves selecting a coordinate $1 \leq k \leq d$ to be split and then the $k$-th interval in the product of rectangle $\mathcal{A}$ undergoes this hierarchical split. A hierarchical split of $\mathcal{A}$ produces two sub-rectangles $\mathcal{A}_1$ and $\mathcal{A}_2$ where $\mathcal{A}_1$ takes the form $\mathcal{A}_1=\Pi_{j=1}^{k-1}[a_j,b_j]\times [a_k,l] \times \Pi_{j=k+1}^d[a_j,b_j]$
for some $1 \leq k \leq d$ and $a_k\leq l \leq b_k$ and $\mathcal{A}_2=\mathcal{A}\cap \mathcal{A}_1^c$. 

\subsection{Intuition on the optimization problem \eqref{optimization}}
To provide intuition on the optimization problem \eqref{optimization} and relate it to the tree-based subgroup analysis, we consider a simple case with $M^* = 2$ subgroups induced by a single hierarchical split of the covariate space. In this scenario, the optimization problem \eqref{optimization} simplifies to: 
\begin{align*}
    \hat{\boldsymbol{\gamma}}_1,\hat{\boldsymbol{\gamma}}_2,\hat{\mathcal{A}}_1,\hat{\mathcal{A}}_2=\arg\min_{ \mathcal{A}_1, \mathcal{A}_2;\boldsymbol{\gamma}_1, \boldsymbol{\gamma}_2\in \mathbb{R}^{2d+2}} L_n\big(\boldsymbol{\gamma}_1,\boldsymbol{\gamma}_2, \mathcal{A}_1,\mathcal{A}_2\big),
\end{align*}
\begin{equation}\label{criteria}
\begin{aligned}
L_n\big(\boldsymbol{\gamma}_1,\boldsymbol{\gamma}_2,  \mathcal{A}_1, \mathcal{A}_2\big) = \sum_{i=1}^n \mathrm{1}_{(\boldsymbol{X}_i\in \mathcal{A}_1)} ||Y_i - \boldsymbol{\gamma}_1^T\boldsymbol{Z}_i||_2^2+\sum_{i=1}^n \mathrm{1}_{(\boldsymbol{X}_i\in \mathcal{A}_2)} ||Y_i - \boldsymbol{\gamma}_2^T\boldsymbol{Z}_i||_2^2.
\end{aligned}
\end{equation}
where $\mathcal{A}_1$ and $\mathcal{A}_2$ are produced by a hierarchical split of the sample covariate space $\mathcal{A}$. This optimization problem is indeed equivalent to performing a single split in the regression tree with the splitting criteria being the goodness-of-fit measure for the outcome regression model $\eta_t$. Thus, solving the optimization problem \eqref{optimization} is equivalent to finding an optimal causal tree that best fits the data and the solution yields a natural estimator of  $\tau(\boldsymbol{X}_i,\Pi^*)$:
$$
\hat{\tau}(\boldsymbol{X}_i,\hat{\Pi})=\sum_{m=1}^{{\hat{M}(\hat{\Pi})}}\mathrm{1}_{(\boldsymbol{X}_i\in \hat{\mathcal{A}}_m)}\cdot(\hat{\mu}_m+\hat{\boldsymbol{\beta}}_m^\top \hat{\bar{\boldsymbol{X}}}_m).
$$

\section{A simple example to illustrate how fusion constraint helps the SATE estimation}\label{app:eg fusion}

This localized approach can hamper both subgroup identification and causal effect estimation. To illustrate this, let us consider a simple example: 
\[
Y = \sum_{m=1}^M l_m(\boldsymbol{\gamma}_m,\boldsymbol{Z}_i) \cdot \mathrm{1}_{\boldsymbol{X} \in \mathcal{A}_m}+\epsilon.
\]  
where \(
l_m(\boldsymbol{\gamma}_m,\boldsymbol{Z}_i)=\mu_m  T + \alpha_{1m} X_1 + \alpha_{2m} X_2 + \beta_{1m} T X_1 + \beta_{2m}  T X_2.\)
Under this model, the SATE for $\mathcal{A}_m$ is:  
\[
\tau (\mathcal{A}_m) = \mu_m + \beta_{1m}  \mathbb{E}[X_1|\boldsymbol{X} \in \mathcal{A}_m] + \beta_{2m} \mathbb{E}[X_2|\boldsymbol{X} \in \mathcal{A}_m].
\]  
We assume some parameters are homogeneous across different subgroups in this model, while others remain heterogeneous. For example, $\beta_{2,m} = \beta_2$ for all $m$. In this scenario, the SATE and the relationship between outcomes and covariates can still vary across subgroups. However, relying solely on local data for estimation will lead to less accurate estimation. To see this, we provide further investigations below:

If the estimated subgroups and parameters are $\hat{\mathcal{A}}_m$ and $\Big(\hat{\mu}_m, \hat{\beta}_{1,m}, \hat{\beta}_{2,m}\Big)$, the SATE is estimated by:  
\[
\hat{\tau} (\hat{\mathcal{A}}_m) = \hat{\mu}_m + \hat{\beta}_{1m} \bar{X}_1(\hat{\mathcal{A}}_m) + \hat{\beta}_{2m} \bar{X}_2(\hat{\mathcal{A}}_m),
\]  
where $ \bar{X}_1(\hat{\mathcal{A}}_m)$ and $ \bar{X}_2(\hat{\mathcal{A}}_m)$ are the sample means within $\hat{\mathcal{A}}_m$. The local estimation nature of the tree algorithm can impair $\hat{\tau} (\hat{\mathcal{A}}_m)$ in two ways: (1) Estimation within individual leaves may involve limited observations, causing overfitting and inaccurate estimates of parameters such as $\hat{\beta}_{2,m}$. (2) Constructing a decision tree involves a dynamic interplay between tree partitioning and local node estimation. Such an overfitting issue in local node parameter estimation can further reduce subgroup identification accuracy. This claim is validated in our simulations presented in Section \ref{sec:simulation}.

\end{document}